\pdfoutput=1

\documentclass[11pt]{article}

\usepackage[final]{acl}

\usepackage{times}
\usepackage{latexsym}

\usepackage[T1]{fontenc}

\usepackage[utf8]{inputenc}

\usepackage{microtype}

\usepackage{inconsolata}

\usepackage{graphicx}

\usepackage[utf8]{inputenc} 
\usepackage[T1]{fontenc}    
\usepackage{hyperref}       
\usepackage{url}            
\usepackage{booktabs}       
\usepackage{amsfonts}       
\usepackage{nicefrac}       
\usepackage{microtype}      
\usepackage{xcolor}         
\usepackage{amsmath}
\usepackage{graphicx}
\usepackage{wrapfig}
\usepackage{multirow}
\usepackage{listings}
\usepackage{caption}
\usepackage{dirtree}
\usepackage{subcaption}
\usepackage{rotating}
\usepackage{enumitem}
\usepackage{algorithmic}
\usepackage[linesnumbered,ruled,vlined]{algorithm2e}
\usepackage{tocloft}
\usepackage{t1enc}
\usepackage{lipsum}
\renewcommand{\DTcomment}[1]{\texttt{\textit{#1}}}

\newtheorem{theorem}{Theorem}
\newtheorem{Observation}[theorem]{Observation}

\title{CollagePrompt: A Benchmark for Budget-Friendly Visual Recognition with GPT-4V}

\author{
 \textbf{Siyu Xu\textsuperscript{1}},
 \textbf{Yunke Wang\textsuperscript{1}},
 \textbf{Daochang Liu\textsuperscript{2}},
 \textbf{Bo Du\textsuperscript{3}},
 \textbf{Chang Xu\textsuperscript{1}\footnotemark[1]},
\\
\\
 \textsuperscript{1}School of Computer Science, The University of Sydney, Australia \\ 
 \textsuperscript{2}School of Physics, Mathematics\&Computing, The University of Western Australia, Australia \\
 \textsuperscript{3}School of Computer Science, Institute of Artificial Intelligence, Wuhan University, China.
\\
 \small{
    \texttt{\{s.xu,c.xu,yunke.wang\}@sydney.edu.au}, \texttt{daochang.liu@uwa.edu.au}, \texttt{dubo@whu.edu.cn}
 }
}

\begin{document}
\maketitle
\renewcommand{\thefootnote}{\fnsymbol{footnote}}
\footnotetext[1]{Corresponding authors.}
\begin{abstract}
Recent advancements in generative AI have suggested that by taking visual prompts, GPT-4V can demonstrate significant proficiency in visual recognition tasks. Despite its impressive capabilities, the financial cost associated with GPT-4V's inference presents a substantial barrier to its wide use. To address this challenge, we propose a budget-friendly collage prompting task that collages multiple images into a single visual prompt and makes GPT-4V perform visual recognition on several images simultaneously, thereby reducing the cost. We collect a \textit{dataset} of various collage prompts to assess its performance in GPT-4V's visual recognition. Our evaluations reveal several key findings: 1) Recognition accuracy varies with different positions in the collage. 2) Grouping images of the same category together leads to better visual recognition results. 3) Incorrect labels often come from adjacent images. These findings highlight the importance of image arrangement within collage prompt. To this end, we construct a \textit{benchmark} called \textbf{CollagePrompt}, which offers a platform for designing collage prompt to achieve more cost-effective visual recognition with GPT-4V. A \textit{baseline} method derived from genetic algorithms to optimize collage layouts is proposed and two \textit{metrics} are introduced to measure the efficiency of the optimized collage prompt. Our benchmark enables researchers to better optimize collage prompts, thus making GPT-4V more cost-effective in visual recognition. The code and data are available at this project page \url{https://collageprompting.github.io/}.
\end{abstract}

\section{Introduction}
\label{sec:intro}
With the rapid development of generative AI, various large language models (LLMs)~\cite{chang2023survey,zhao2023survey,thirunavukarasu2023large} have emerged as generative tools. Beyond text, these models have expanded their capabilities to include text-to-image generation such as Stable Diffusion~\cite{rombach2022high}, and text-to-video generation, as seen in Sora~\cite{sora2023texttovideo}. ChatGPT~\cite{brown2020language}, as the most well-known LLMs, has shown it can have natural and coherent conversations, making it a powerful tool in daily life and different industry fields. As the latest version of ChatGPT, GPT-4V is a multi-modal LLM capable of processing both text and images. This capability allows it to be applied to a wider range of applications and tasks. There are many technical reports and user studies~\cite{li2023comprehensive,lin2023mm,shi2023exploring,wen2023road,yang2023dawn,zhou2023exploring} about GPT-4V, which conducted thorough evaluations of its capabilities from various aspects. 

In \cite{wu2023gpt4vis}, the visual capabilities of GPT-4V are investigated within the framework of zero-shot visual recognition tasks, such as image and video recognition. The evaluation of visual capabilities is quite straightforward: images and candidate categories are directly fed into GPT-4V for relevance ranking, yielding Top-1 and Top-5 prediction results. Video and point cloud data are uniformly sampled to generate a set of images, which are then processed by GPT-4V for visual recognition. GPT-4V has achieved remarkable performance across various visual recognition tasks, surpassing previous customized solutions\cite{wang2018learning,dosovitskiy2020image}. 
However, its financial cost associated with its inference can be significant. Specifically, performing image recognition on the ImageNet-1K dataset ~\cite{russakovsky2015imagenet} requires approximately \$1 for every 20 images, leading to a total evaluation cost of over \$2,500 for the entire dataset. If we consider the rate limits of maximum API requests per minute, the costs will be even higher. Thus, employing GPT-4V for visual recognition tasks is expensive, and it is meaningful to adopt a more budget-friendly way for GPT-4V's visual recognition.
\begin{figure*}[t]
\centering
    \includegraphics[width=0.99\textwidth]{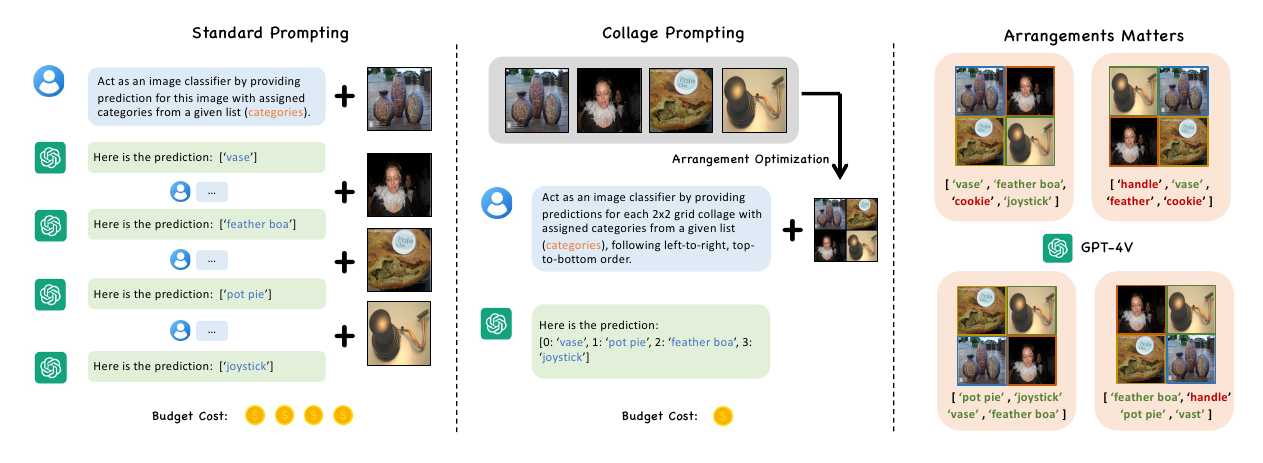}
    \vskip -0.05in
    \caption{Visual recognition of GPT-4V with different prompting ways. (a): \textit{Standard Prompt} takes one image as visual prompt for each GPT-4V's run. (b): \textit{Collage Prompt} concatenates multiple images into one visual prompt and predicts class for all images in each inference. (c): The arrangement of images within collage prompting leads to significantly different results. \textcolor{green}{Green} indicates an accurate prediction while \textcolor{red}{red} indicates an wrong prediction.}
    \label{fig:motivation}
    \vskip -0.15in
\end{figure*}

In GPT-4V's visual recognition, only an image is used as the visual prompt. This standard prompting way fails to fully release the potential capacity of GPT-4V, which is able to process multiple inquiries within one prompt simultaneously. Motivated by this idea, we propose Collage Prompting, a new task of prompting for GPT-4's visual recognition for budget-friendly inference. In collage prompt, multiple images are collaged into one visual prompt with equal size. GPT-4V is then requested to recognize the class for all images within this prompt. The overall process is shown in Figure \ref{fig:motivation}. Since collage prompting allows for the recognition of multiple images in one run of GPT-4V, it significantly reduces the average cost of visual recognition. 

\noindent\textbf{Benchmark.} 
Based on the observation that different arrangements in the collage prompt could lead to rather large variance of accuracy in GPT-4V's recognition, we further construct a benchmark for arrangement optimization. The benchmark collects a comprehensive dataset that contains various collage prompts, which is then used to either assess the performance of collage prompts or provide a platform to develop algorithms that can optimize collage prompts for more cost-effective recognition. Based on the idea of genetic algorithm, we develop a baseline \textit{Learn to Collage} (LCP) to optimize the arrangement. In our baseline, the collage prompt is represented as a graph, and a collage predictor is used to estimate the expected accuracy of this collage prompt. LCP is then used to search for the best arrangement in several iterations. Two new metrics are proposed to measure the cost-effectiveness of the developed algorithm.

\noindent\textbf{Contributions.} We make three contributions in this paper. First, we propose a budget-friendly prompting approach for GPT-4V. By involving multiple images into a single visual prompt, GPT-4V can process multiple images in one inference run, thus reducing the overall expense greatly. Second, we collect the benchmark dataset of collage prompt. The datasets contain various collage prompts from the ImageNet-1K training set and their accuracy in GPT-4V's image recognition. This dataset is meaningful for studying the effectiveness of collage prompting. Third, we propose a genetic algorithm-based optimization method for collage arrangement. This approach aims to optimize the arrangement for collage prompts and improve image recognition accuracy within GPT-4V.

\section{Related Works}

\noindent\textbf{Exploration of GPT-4V.}
The state-of-the-art large multi-modal model GPT-4V was firstly launched at September 2023 and has demonstrated its strong visual capability in different fields. Early works~\cite{wu2023early,yang2023dawn} conducted a user study of GPT-4V, where operations were completed by entering prompts on a web interface for GPT-4V. Furthermore, the release of GPT-4V API in November 2023 opened up new opportunities for both academic community and industry to thoroughly evaluate GPT-4V's performance across various visual benchmarks and provide quantitative data beyond what user studies can offer. GPT-4V's capacities in multimodal medical diagnosis has been explored in~\cite{wu2023can,yang2023performance,deng2024vision}, where GPT-4V can process different imaging modalities like CT and MRI in medical scene. GPT-4V can also be utilized to operate robots from providing instructions by taking multimodal input in autonomous driving~\cite{cui2024survey,han2024dme,wen2023road} and task planning~\cite{wake2023gpt,wang2024large,hu2023look}. Additionally, GPT-4V has been widely used in advancing video understanding~\cite{lin2023mm}, conducting OCR recognition~\cite{shi2023exploring}, acting as an intelligent web agent~\cite{zheng2024gpt}, and dealing with each observation data~\cite{zhang2024good}. ~\cite{wu2023gpt4vis} is the first work that considers to adopt extensive quantitative analysis utilizing the established visual benchmarks. However, the evaluation of GPT-4V on visual benchmarks could lead to large expense and it is important to adopt a budget-friendly inference scheme for the evaluation of GPT-4V.

\noindent\textbf{Prompt Engineering in LLMs.}
Prompt engineering has emerged as a crucial technique for unlocking the potential of pre-trained large language models (LLMs) and vision-language models (VLMs). The concept of prompt engineering was initially explored and popularized in the LLMs~\cite{liu2023pre,tonmoy2024comprehensive,chen2023unleashing} and VLMs~\cite{wu2023visual,bahng2022exploring}. The most common prompting way is zero-shot prompting~\cite{radford2019language,cheng2023batch}, which offers a paradigm shift in leveraging large LLMs. This method significantly reduces the dependency on vast amounts of training data by employing strategically formulated prompts to steer the model towards executing new tasks. While the primary focus in the field has been on creating prompts that can release the potential capacities of LLMs, we focus on developing a prompting approach that prioritizes cost-efficiency.

\begin{figure*}[t]
  \centering
  \begin{subfigure}{0.78\linewidth}
  \includegraphics[width=1\textwidth]{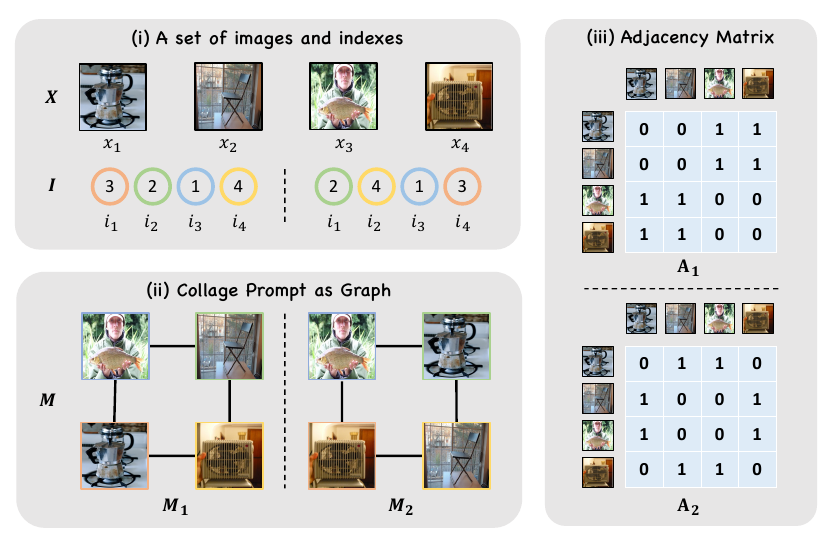}
    \caption{}
    \label{fig:graph}
  \end{subfigure}
  \hfill
  \begin{subfigure}{0.21\linewidth}
    \includegraphics[width=1\textwidth]{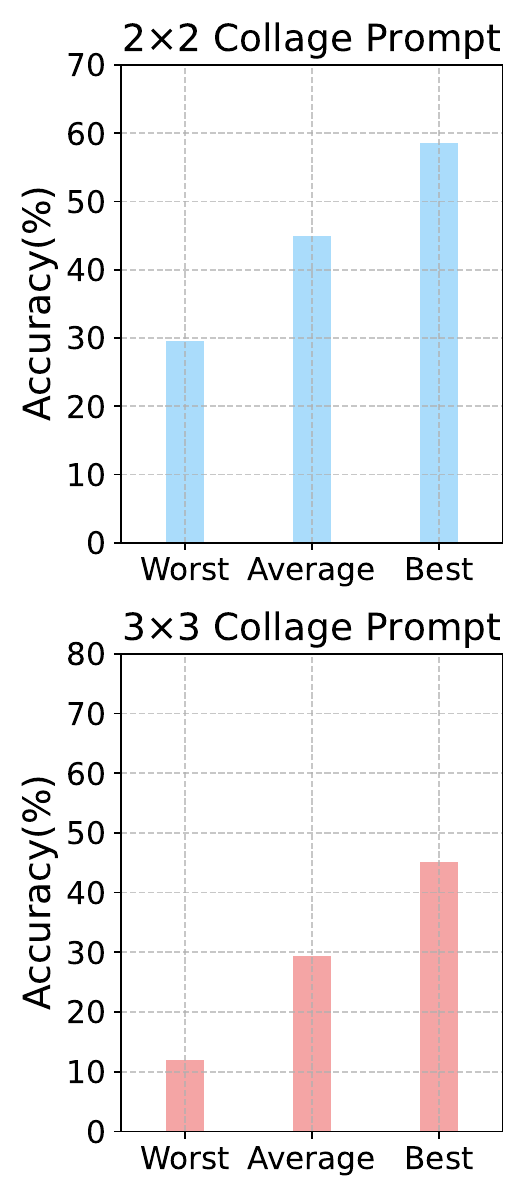}
    \caption{}
    \label{fig:short-b}
  \end{subfigure}
   \vskip -0.10in
  \caption{\textbf{(a):} The workflow of forming the collage prompt from a set of images and related indexes. For a set of images $\textbf{X}$ with two different position indexes $\textbf{I}$, we can obtain two collage prompts $\textbf{M}_1$ and $\textbf{M}_2$. Regarding $\textbf{X}$ as the node of a graph, the adjacency matrix of $\textbf{M}_1$ and $\textbf{M}_2$ can be represented as $\textbf{A}_1$ and $\textbf{A}_2$. \textbf{(b):} The average accuracy of collage prompts within evaluation datasets using the `Worst,' `Average', and `Best' arrangement.}
  \vskip -0.15in
  \label{fig:short}
\end{figure*}

\section{Collage Prompting}

GPT-4V has enabled us to perform comprehensive visual recognition. However, each inference performed by GPT-4V incurs a financial cost, which is determined by the number and type of input and generated tokens\footnote{https://openai.com/pricing, based on pricing as of March 1, 2024.}. Specifically, for image recognition tasks involving images of $512\times512$ resolution, approximately 5000 tokens are consumed per image. Standard prompting of GPT-4V involves presenting a single image as a visual prompt and processing each image in the dataset individually. With this standard prompting, the expense of evaluating a dataset with 10,000 images could exceed \$500. This method is costly, and a more budget-friendly approach is to process multiple images simultaneously in a single inference run.

Motivated by this idea, we propose Collage Prompting, an efficient alternative to standard visual prompting. Collage Prompting involves concatenating multiple images into a single visual prompt, allowing for simultaneous processing in a single inference run. For example, employing a nine-grid collage prompt can decrease expenses to just $1/9$ of what standard individual image prompting incurs. Moreover, collage prompt not only significantly reduces costs but also processes multiple images with a single API request, thereby reducing server load and inference time. This approach is particularly beneficial for large-scale, high-frequency applications of multimodal foundational models, such as using GPT-4V for image captioning or employing GPT-4o for reading video streams.

\noindent\textbf{Preliminary of collage prompt.}
By assembling $K$ images into one visual prompt, the collage prompt $\textbf{M}$ is designed to be a $\sqrt{K}\times\sqrt{K}$ grid and each grid contains one image. For example, a collage of four images might be presented in a quadrant grid, while nine images could be arranged in a nine-grid format. 
Supposing we have a set of $K$ images $\textbf{X}=[x_1,x_2,...,x_K]$ and its related position indexes $\textbf{I}=[i_1,i_2,...,i_K]$, where $i_j$ indicates the position number of image $x_j$ in the collage prompt, starting from $0$ in the top left corner to $K-1$ in the bottom right corner. Hence, the row position $r$ and column position $c$ of image $x_j$ can be specified as $r=\lfloor (i_j-1)/\sqrt{K} \rfloor$ and $c=(i_j-1)\mod\sqrt{K}$ respectively. While collage prompt $\textbf{M}$ has the same size as the standard prompt, GPT-4V can thus take this collage prompt as input and generate the predicted class for all images within the collage prompt. 
Regarding the collage prompt $\textbf{M}$ as graph $\textbf{M}=(\textbf{A},\textbf{F})$ with $K$ nodes, each node $f_i\in\mathbb{R}^l$ denotes the feature of $x_i$. The adjacency matrix $\textbf{A}\in\mathbb{R}^{K\times K}$ suggests the relative positions of $\textbf{X}$, where two adjacent images are considered to have an undirected edge. For two images $x_p$ and $x_q$ that satisfies either $|c_p-c_q|=1$ when $r_p=r_q$ or $|r_p-r_q|=1$ when $c_p=c_q$, we consider these two images have an edge and set $\textbf{A}[p,q]=\textbf{A}[q,p]=1$. The workflow of representing the collage prompt as a graph is illustrated in Figure \ref{fig:graph}. 

\noindent\textbf{Arrangements matter.}
To fill multiple images into the collage prompt, the arrangement of these images could be various. As shown in Figure \ref{fig:graph}, by setting different $\textbf{I}$, the arrangement of images within the collage prompt is different. Technically, a quadrant-grid collage has $24(4!)$ potential image arrangements, whereas a nine-grid format can exceed $360,000(9!)$ possibilities. Our findings in Figure \ref{fig:short-b} indicate that different arrangements can yield varying levels of accuracy, underscoring the importance of image arrangement. Therefore, we hope to find better arrangement within the collage to minimize the accuracy loss of GPT-4V.

\section{A Benchmark of Collage Prompting}
The arrangement of images within a collage prompt significantly impacts the overall recognition accuracy of GPT-4V. For any given set of images forming a collage, there should exist one or more optimal arrangements that maximize overall recognition accuracy. Thus, we conduct a benchmark study of collage prompting with two primary objectives: \textbf{1)} to study the effect of different arrangements on GPT-4V’s recognition accuracy and \textbf{2)} to provide a benchmark for developing algorithms to optimize collage arrangements. 
In this section, we first present a comprehensive collage prompt dataset to assess the performance of various collage prompts. Three key observations suggest that there is a need to conduct arrangement optimization. Then, we propose a baseline method to learn the layout of the collage and two metrics that reflect the cost-effective trait of GPT-4V's visual recognition are proposed. 

\begin{figure*}[!tbp]
    \centering
    \begin{minipage}[b]{1\linewidth}
        \begin{subfigure}[b]{0.24\textwidth}
            \centering
            \includegraphics[width=\textwidth]{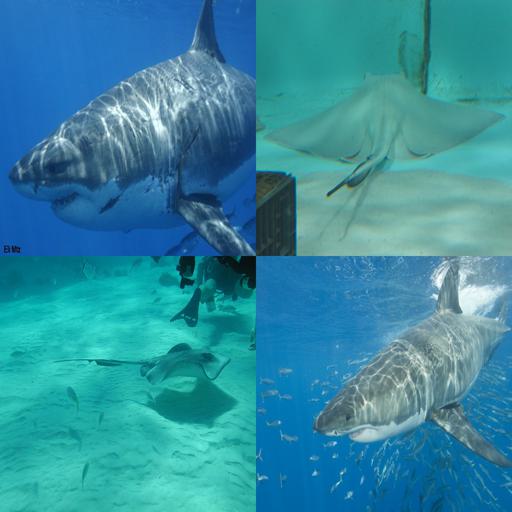}
            \caption{"0: ‘great white shark’, 1: ‘stingray’, 2: ‘{\color{red}great white shark}’, 3: ‘great white shark’"}
            \label{fig:ob:2a}
        \end{subfigure}
        \hfill
        \begin{subfigure}[b]{0.24\textwidth}
            \centering
            \includegraphics[width=\textwidth]{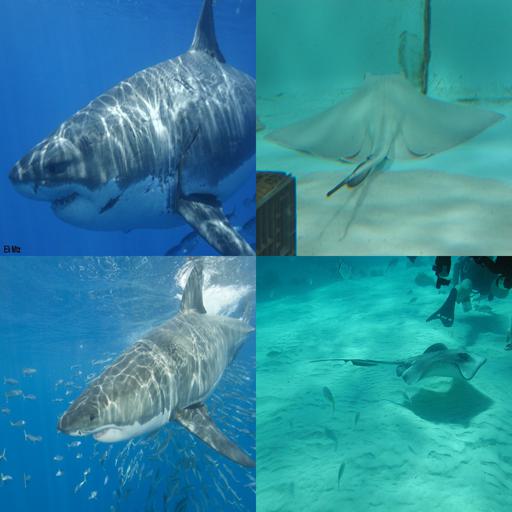}
            \caption{"0: ‘great white shark’, 1: ‘stingray’, 2: ‘great white shark’, 3: ‘stingray’"}
            \label{fig:ob:2b}
        \end{subfigure}
        \hfill
        \begin{subfigure}[b]{0.24\textwidth}
            \centering
            \includegraphics[width=\textwidth]{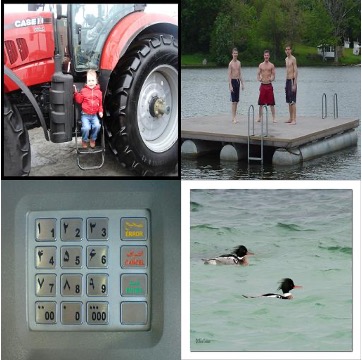}
            \caption{"0: ‘tractor’, 1: ‘{\color{red}automated teller machine}’, 2: ‘{\color{red}red-breasted merganser}’, 3: ‘{\color{red}dive}’"}
            \label{fig:ob:3a}
        \end{subfigure}
        \hfill
        \begin{subfigure}[b]{0.24\textwidth}
            \centering
            \includegraphics[width=\textwidth]{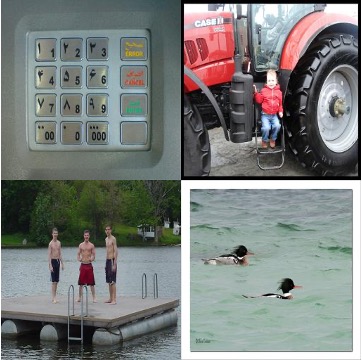}
            \caption{"0: ‘automated teller machine’, 1: ‘tractor’, 2: ‘{\color{red}dock}’, 3: ‘red-breasted merganser’"}
            \label{fig:ob:3b}
        \end{subfigure}
        \caption{
        (a) and (b) demonstrate the effect of \textbf{category clustering}, where placing images of the same category together increases overall recognition accuracy. (c) and (d) illustrate \textbf{localization errors}, where GPT-4V predicts the correct labels but outputs them to incorrect positions in the collage.
        }
        \label{fig:observation}
    \end{minipage}
    \vspace{-1.6em}
\vskip -0.15in
\end{figure*}

\begin{figure}[!tbp]
    \centering
    \includegraphics[width=3.0in]{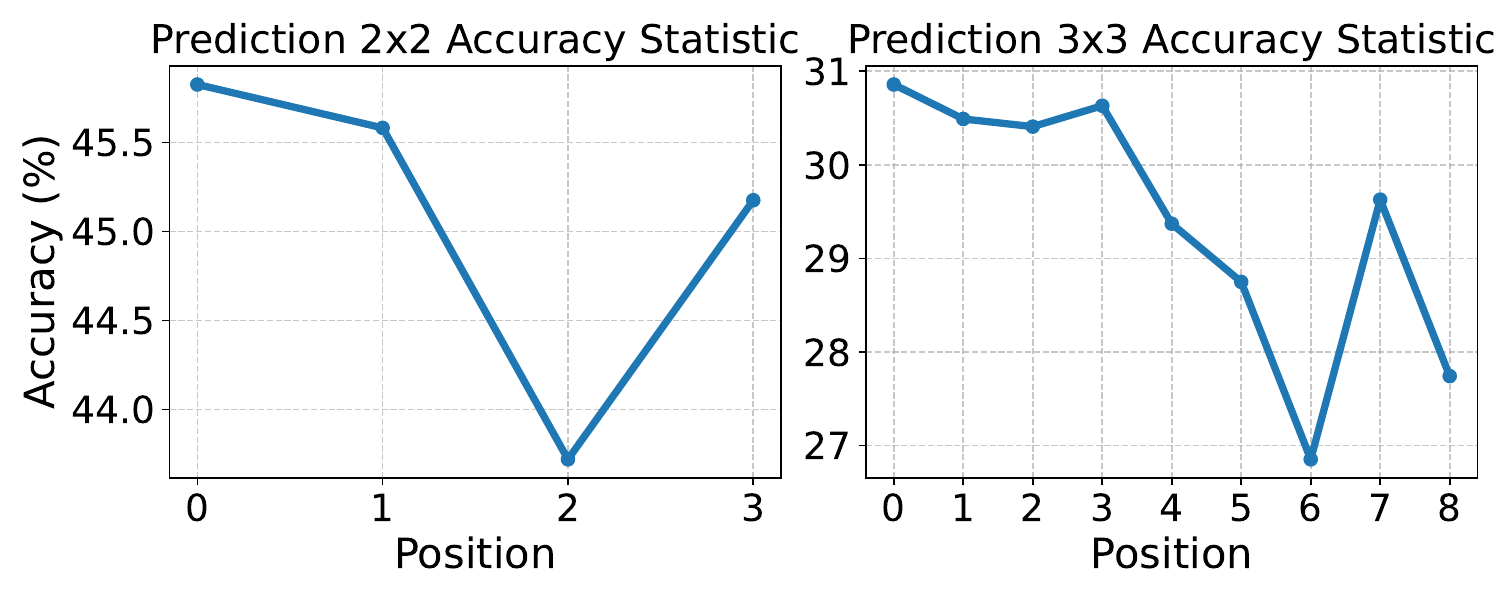}
    \vskip -0.1in
    \caption{Average Prediction Accuracy by Position.}
        \vskip -0.2in
    \label{fig:pos_stat_wrap}
\end{figure}

\subsection{Dataset}
To construct a collage prompting dataset with various arrangements, we generate different collages (\textit{i.e.}, \textbf{A}) for the same set of images (\textit{i.e.}, \textbf{X}). 
We first uniformly sample a sub-dataset that contains 100,000 images from the training set of ImageNet-1K. This subset is then divided into $L$ groups, and each group contains $K$ images. By executing $p$ random shuffles of the images within each group, we generated a collection of $L \times p$ unique collage prompts. We collect two collage prompting datasets with a quadrant-grid collage prompt and a nine-grid collage prompt. 
For the quadrant-grid collage prompt, $L$ is set to be 25,000 and $p$ is set to be 5. For the nine-grid collage prompt, $L$ is set to be 11,111 and $p$ is set to be 10. These collage prompts are then sent into GPT-4V's API for image recognition and the accuracy $y$ of each prompt can be obtained. The final dataset $\mathcal{D}$ thus includes pairs $\{\textbf{M}_i, y_i\}$ for each of the $L \times p$ prompts, providing a comprehensive basis for analyzing the effectiveness of different collage configurations in GPT-4V's visual recognition. 
We conducted an in-depth analysis of the collected collage prompting datasets and observed the following patterns.

\begin{Observation}[Position Accuracy Variance]
    Different positions within the collage grid have varying accuracy in GPT-4V's visual recognition. 
\end{Observation}
As shown in Figure \ref{fig:pos_stat_wrap}, the top-left position in both $2\times2$ and $3\times3$ grids tends to have the highest accuracy, with accuracy decreasing towards the center and bottom-left positions, which have the lowest accuracy. Accuracy then improves slightly for the last row. This pattern suggests potential model fatigue when processing central images in the collage, leading to lower accuracy that recovers as the model approaches the final row. Based on this observation, a natural idea to optimize the arrangement is to place `hard' images into positions with higher accuracy while leaving `easy' images to remaining positions.

\begin{Observation}[Category Clustering]
    Placing images of the same class together in a collage improves accuracy, while pushing images of the same class away from each other degrades the accuracy.
\end{Observation}
We observed that in both $2\times2$ and $3\times3$ collages, placing images of the same class together significantly improves GPT-4V's overall recognition accuracy. Conversely, when the order is shuffled and images of the same class are not adjacent, the accuracy decreases. As illustrated in Figure \ref{fig:ob:2a} and \ref{fig:ob:2b}, GPT-4V predicts one of the stingrays incorrectly when the great white shark and stingray are on separate diagonals in the collage, and correctly if the same class is adjacent. This improvement can be due to clustering images of the same class reduces the complexity of batch recognition for GPT-4V. 

\begin{Observation}[Localization Errors]
    GPT-4V often makes localization errors, predicting labels for adjacent images incorrectly. 
\end{Observation}
We analyzed the prediction errors in $2\times2$ and $3\times3$ collages and found that the incorrectly predicted labels often correspond to images in adjacent positions. This indicates that the model correctly identifies the images but outputs the predictions to the wrong locations due to localization inaccuracies. For instance, in Figure \ref{fig:ob:3a} and \ref{fig:ob:3b}, GPT-4V predicts the `automated teller' machine and `red-breasted merganser', but outputs to the wrong positions in the collage. When the arrangement order is changed, the model outputs the correctly predicted labels to the correct positions.

\begin{figure*}[!t]
    \centering
    \includegraphics[width=5.5in]{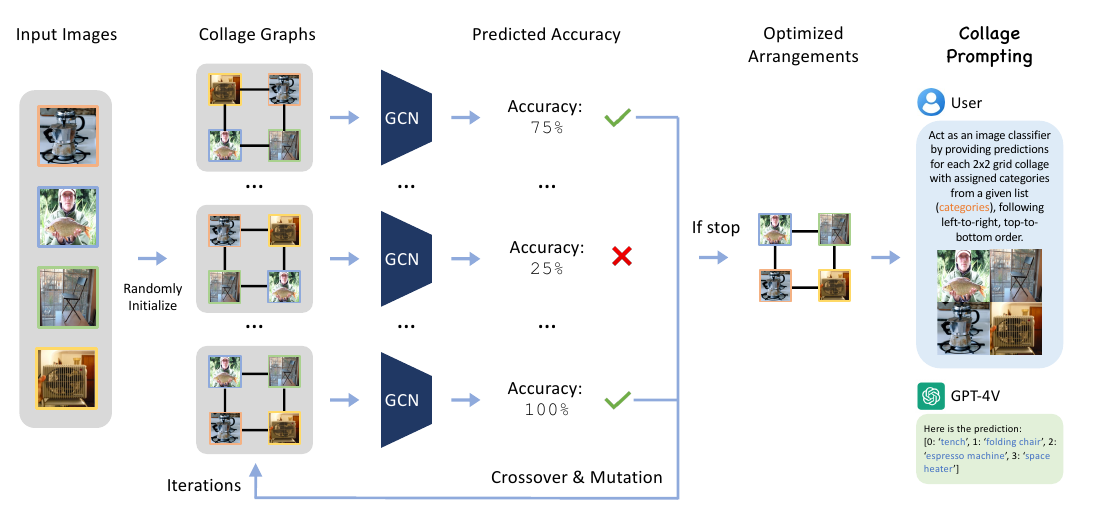}
    \caption{An overview of baseline method LCP. Starting with a set of images, index sets are randomly initialized, which forms multiple collage graphs. After predicting the accuracy of each collage graph via $G_{\theta^\ast}$, collage graphs that achieve top-$T$ accuracy are selected for crossover and mutation operations. This iterative process continues until reaching the maximum specified iteration and we can obtain the optimized arrangements.
    }
    \label{fig:inference}
    \vskip -0.1in
\end{figure*}

\noindent\textbf{Visual Bias Analysis}
Our analysis of collage prompting reveals that GPT-4V exhibits biases in visual recognition based on image placement. Specifically, (1) \textit{Position Accuracy Variance} accuracy varies across collage positions, with top-left performing best and central positions showing lower accuracy, likely due to model fatigue. (2) \textit{Category Clustering} grouping similar images enhances accuracy, while separating them reduces it. (3) \textit{Localization Errors} labels are often misassigned to adjacent images, indicating spatial misalignment challenges. These insights highlight the need for optimized arrangements to mitigate biases. Our baseline algorithm leverages these biases to enhance collage layouts, improving recognition accuracy while maintaining cost efficiency. For further details and visual examples, see Appendix~\ref{app:visual_analysis}.

\subsection{Baseline: Learning to Collage}
The above observations from the collage prompting dataset further highlights the importance of arrangement optimization within the collage prompt. To solve this task, we propose a baseline method Learning to Collage (LCP) in the benchmark. This baseline method involves two processes: training a collage predictor for accuracy prediction and refining collage's arrangement via genetic algorithm.

\noindent\textbf{Training of Collage Predictor.} \quad
As stated in the previous subsection, collage prompt $\textbf{M}$ is represented as a graph. To predict the performance of these collage prompts, we employ GCN~\cite{zhang2019hierarchical}, denoted as $G_\theta(\textbf{A},\textbf{F})$, to process the graph data and predicts the expected accuracy of the collage prompt. 

Given the evaluation dataset $\mathcal{D}=\{\textbf{M}_i, y_i\}_{i=1}^{L\cdot p}$, the update of the $G_\theta(\textbf{A},\textbf{F})$ at $k$-th iteration can be expressed as,
\begin{equation}
\small
    \theta_{k+1}=\theta_k- \frac{\eta}{b} \sum_{i=0}^{b} \nabla_\theta L(G_\theta(\textbf{A}_i,\textbf{F}_i),y_i),
    \label{eq32}
\end{equation}
where $b$ is the batch size of training, $\eta$ is the learning rate and $L$ denotes the MSE loss. At the convergence step, $\theta^\ast$ will be obtained and $\textbf{G}_{\theta^\ast}$ can be used to indicate the expected accuracy of the collage prompt $\textbf{M}$.

\begin{table*}[!t]
\small
    \centering
        \caption{Benchmark results of collage prompting with different sizes.}
        \vskip 0.05in
\setlength{\tabcolsep}{3.2mm}{
    \begin{tabular}{llrcccccc}
    \toprule
     & \multirow{2}{*}{Datasets} & \multirow{2}{*}{Cost(\$/1k)} & \multicolumn{2}{c}{Top-1 Accuracy} & \multicolumn{2}{c}{CER} & \multicolumn{2}{c}{PCE} \\
     \cmidrule(lr){4-5}\cmidrule(lr){6-7}\cmidrule(lr){8-9}
        &  &  & Random & Baseline &  Random & Baseline & Random & Baseline \\
    \midrule
    
        \multirow{11}{*}{\begin{turn}{90}\textbf{2}$\times$\textbf{2}\end{turn}}
          &ImageNet-1K  &  \$12.83 	 & 	39.4\%	 & 	\textbf{45.7\%}	 & 	4.99	 & \textbf{	5.38}		 & 	5.49	 & 	\textbf{10.60 }\\
          &Caltech101  &  \$1.81 	 & 	88.4\%	 & 	\textbf{90.8\%}	 & 	13.37	 & 	\textbf{13.52}		 & 	2.15	 & 	\textbf{3.18 }\\
          &OxfordPets  &  \$1.25 	 & 	70.1\%	 & 	\textbf{71.8\%}	 & 	13.20	 & 	\textbf{13.34}		 & 	1.18	 & 	\textbf{1.20 }\\
          &StanfordCars  &  \$5.28 	 & 	29.8\%	 & \textbf{	32.0\%}	 & 	5.65	 & \textbf{	5.88}		 & 	1.74	 & 	\textbf{1.83} \\
          &Flowers102  &  \$2.15 	 & 	49.8\%	 & 	\textbf{51.1\%}	 & 	9.75	 & 	\textbf{9.88}		 & 	1.36	 & 	\textbf{1.39} \\
          &Food101  &  \$2.02 	 & 	62.2\%	 & \textbf{	64.2\%}	 & 	11.05	 & 	\textbf{11.21}		 & 	1.40	 & 	\textbf{1.46} \\
          &Aircraft  &  \$2.15 	 & 	\textbf{18.5\%}	 & 	17.7\%	 & 	\textbf{5.57}	 & 	5.42		 & 	1.45	 & 	\textbf{1.42} \\
          &SUN397  &  \$5.39 	 & 	46.5\%	 & \textbf{	48.7\%}	 & 	7.12	 & 	\textbf{7.29}		 & 	4.23	 & 	\textbf{6.02} \\
        &DTD  &  \$1.27 	 & 	48.6\%	 & \textbf{	52.0\%}	 & 	11.02	 & 	\textbf{11.40}		 & 	1.44	 & 	\textbf{1.71} \\
      & EuroSAT &  \$0.86 	 & 	42.9\%	 & 	\textbf{53.4\%}	 & 	11.21	 & 	\textbf{12.52}		 & 	\textbf{1.47}	 & 	1.16 \\
        &UCF101  &  \$2.06 	 & 	55.2\%	 & 	\textbf{58.1\%}	 & 	10.39	 & 	\textbf{10.65}		 & 	1.26	 & 	\textbf{1.30}\\
         \midrule
        \multirow{11}{*}{\begin{turn}{90}\textbf{3}$\times$\textbf{3}\end{turn}}
          &ImageNet-1K  &  \$5.70 	&	28.1\%	&	\textbf{33.7\%}	&	5.33	&	\textbf{5.90}		&	3.84	&	\textbf{5.01 }\\
          &Caltech101  & \$0.80 	&	79.1\%	&	\textbf{85.4\%}	&	15.26	&	\textbf{15.79}		&	1.48&		\textbf{1.89}\\
          &OxfordPets  &  \$0.55 	&	53.2\%	&	\textbf{59.5\%}	&	13.45	&	\textbf{14.20}		&	1.12&		\textbf{1.14} \\
          &StanfordCars  &  \$2.34 	&	11.6\%	&	\textbf{14.7\%}	&	3.96	&	\textbf{4.68}		&	1.49&		\textbf{1.54} \\
          &Flowers102  &  \$0.95 	&	38.4\%	&	\textbf{43.8\%}	&	10.38	&	\textbf{11.12}			&1.27	&	\textbf{1.33} \\
          &Food101  &  \$0.90 	&	39.9\%	&	\textbf{46.8\%}	&	10.71	&	\textbf{11.63}		&	1.20&		\textbf{1.24} \\
          &Aircraft  & \$0.96 	&	7.0\%	&	\textbf{10.3\%}	&	3.32	&	\textbf{4.52}		&	1.30&		\textbf{1.35} \\
          &SUN397  &  \$2.39 	&	27.6\%	&	\textbf{36.3\%}	&	6.89	&	\textbf{8.03}		&	1.89&		\textbf{2.45} \\
        &DTD  &  \$0.56 	&	37.5\%		&\textbf{44.1\%}	&	11.22	&	\textbf{12.21}		&	1.23	&	\textbf{1.35} \\
      & EuroSAT &  \$0.38 	&	30.4\%	&	\textbf{39.7\%}	&	10.50	&	\textbf{12.14}		&	1.70	&	\textbf{2.40}\\
        &UCF101  &  \$0.91 	&	37.9\%	&	\textbf{44.0\%}	&	10.39	&	\textbf{11.24}		&	1.18	&	\textbf{1.21} \\
        \bottomrule
    \end{tabular}
}
\vskip -0.15in
    \label{tab:cost_acc}
\end{table*}

\begin{table}[!ht]
\centering
  \vspace{-1pt} 
  \begin{minipage}{0.45\textwidth}
    \caption{Results of GPT-4V's zero-shot visual recognition in 11 various datasets\cite{wu2023gpt4vis}.}
    \label{tab:money_cost_grid_cnn_vit}
    \resizebox{\textwidth}{!}{
      \begin{tabular}{@{}lccc@{}}
        \toprule
        Dataset & Cost(\$/1k) & Top-1 Acc. & CER \\
        \midrule
        ImageNet-1K & \$51.30 & 62.0\% & 4.30 \\
        Caltech101 & \$7.24 & 95.5\% & 9.08 \\
        OxfordPets & \$4.99 & 92.6\% & 10.09 \\
        StanfordCars & \$21.10 & 58.3\% & 5.26 \\
        Flowers102 & \$8.58 & 70.6\% & 7.52 \\
        Food101 & \$8.09 & 80.1\% & 8.12 \\
        Aircraft & \$8.61 & 36.0\% & 5.37 \\
        SUN397 & \$21.55 & 57.7\% & 5.20 \\
        DTD & \$5.07 & 59.1\% & 8.17 \\
        EuroSAT & \$3.45 & 36.2\% & 7.19 \\
        UCF101 & \$8.22 & 81.6\% & 8.14 \\
        \bottomrule
      \end{tabular}
    }
  \end{minipage}
  \vskip -0.2in
\end{table}
\noindent\textbf{Arrangement optimization.} \quad
With the trained predictor $\mathbf{G}_{\theta^*}$, we can estimate the accuracy for various arrangements in the collage prompt, which enables the selection of the most effective arrangement to enhance recognition performance with GPT-4V. Due to the vast number of potential arrangements, it is impractical to evaluate each one to identify the optimal arrangement. To efficiently search for the best arrangement within a maximum number of iterations, we use genetic algorithm (GA) that has been widely used for non-differentiable optimization problems \cite{wang2018towards,wang2019evolutionary} to achieve effectively searching.
Following the idea of GA, LCP alternately evaluates the quality of arrangements in the current population and searches for the optimal collage arrangement through operations such as selection, crossover, and mutation. 
As shown in Figure \ref{fig:inference}, LCP consists of several key stages:
\begin{itemize}[left=0pt]
    \item \noindent\textbf{Initialization.} In the initialization phase, for a given set of image features $\mathbf{F}$, we randomly generate a set of position index set $\mathcal{I}=\{{\textbf{I}}_1, {\textbf{I}}_2,...,{\textbf{I}}_P\}$ with the related adjacency matrices $\mathcal{A} = \{\mathbf{A}_1, \mathbf{A}_2, \ldots, \mathbf{A}_P\}$. These matrices represent different possible arrangements of the collage prompt. 
    \item \noindent\textbf{Evaluation.} Subsequently, for each adjacency matrix $\mathbf{A}_i$ in $\mathcal{A}$, we predict its expected accuracy using $\hat{y}_i = G_{\theta^*}(\mathbf{A}_i, \mathbf{F})$, resulting in a set of predicted accuracy $\mathcal{Y} = \{\hat{y}_0, \hat{y}_1, \ldots, \hat{y}_P\}$.
    \item \noindent\textbf{Selection.} During the selection phase, we choose a subset $\tilde{\mathcal{I}}$ of arrangements from $\mathcal{I}$ that correspond to the top-$T$ accuracy in $\mathcal{Y}$, indicating the most promising arrangements, which are then preserved in the next iteration. 
    \item \noindent\textbf{Crossover \& Mutation.} To generate the next generation of arrangements, crossover and mutation operations are applied to the selected subset $\tilde{\mathcal{I}}$. We randomly select two arrangements from $\tilde{\mathcal{I}}$ for crossover and mutation, generating new arrangements for the next iteration. Specifically, in the crossover process, we divide two position indexes $\textbf{I}\in \tilde{\mathcal{I}}$ into segments and cross a segment between them to generate two new position indexes. We retain the new position index with higher expected accuracy. To promote diversity, we randomly select a position index from $\tilde{\mathcal{I}}$ and mutate a randomly chosen segment of the position index. This iterative process continues, refining the search for an arrangement that maximizes the accuracy of collage prompt recognition by GPT-4V.
\end{itemize}

\begin{table*}[!t]
  \caption{Evaluation Cost of different methods in ImageNet-1K.
  }

  \label{tab:money_cost_grid_cnn_vit}
  \centering
  \setlength{\tabcolsep}{3.5mm}{
  \begin{tabular}{@{}lccccccc@{}}
    \toprule
     & Epochs & Accuracy & Training Cost & Test Cost& Total Cost & CER \\
    \midrule
    ViT-B/16 & 300 & 84.53\% & \$3,876.69 & \$32.77 & \$3,909.46 & 0.022\\
    ResNet-50 & 200 & 79.04\% & \$3,063.99 & \$27.64 & \$2,091.63 &  0.038\\
    \midrule
    $1\times1$ Grid  &  - & 62.0\% & - & \$641.25 & \$641.25 & 0.097\\
    $2\times2$ Grid &  - & 45.7\% & \$9.9 & \$160.37 & \$170.27 & 0.268\\
    $3\times3$ Grid &  - & 33.7\% & \$16.5 & \$ 71.25 & \$87.75 & \textbf{0.384}\\
  \bottomrule
  \end{tabular}
  }
\vskip -0.2in
\end{table*}

By iteratively employing these steps, the initial arrangements are updated efficiently until the maximum iterations are achieved. After obtaining the arrangement with the best-expected accuracy, we can apply this arrangement configuration to enhance the performance of collage prompts. Algorithm \ref{app:algorithm_lcp} outlines the steps of the LCP algorithm, while Appendix~\ref{app:lcp_details} provides more details of the baseline method. To support reproducibility, we have released the dataset, baseline code, and model weights on our project page.

\subsection{Metrics}
\label{sec:metrics}
To effectively assess the performance of our collage prompting approach, we utilize \textbf{cost} (\textit{i.e.,} $C_{n\times n}$) and \textbf{accuracy} (\textit{i.e.,} $A_{n\times n}$) as primary evaluation metrics, where $n\times n$ denotes the size of collage prompt. Collage prompt can significantly reduce costs but often at the expense of recognition performance loss. Besides evaluating cost and accuracy separately, we introduce two new metrics for a more comprehensive analysis as follow:

    \noindent\textbf{Cost-Effective Ratio (CER)}: CER applies logarithmic transformations on both primary metrics to improve the distinction and manageability of the values. It is formulated as,
    \begin{equation}
    \small
    \mathrm{CER} =  \frac{\left(\log(A_{n\times n} + 1)\right)^\gamma}{ \log(C_{n\times n} + e)},
    \end{equation}
    
    \noindent\textbf{Precision-Cost Efficiency (PCE)}: PCE signifies the cost saved per accuracy loss, which is formulated as,
    \begin{equation}
    \small
     \mathrm{PCE} = e^{\left(\frac{\lvert C_{n\times n} - C_{1\times 1}\rvert }{\lvert A_{n\times n} - A_{1\times 1}\rvert}\right)}.
    \end{equation}
    
These two metrics balance the trade-offs between accuracy and cost, providing deeper insights into the efficiency of various collage configurations. We also use these metrics to evaluate the performance of our baseline algorithm.

\section{Experiment}
In this section, we benchmarked GPT-4V's zero-shot collage prompt recognition performance on ImageNet-1K~\cite{russakovsky2015imagenet} and 10 other datasets (\textit{e.g.}, Caltech101~\cite{fei2004learning}, OxfordPets~\cite{parkhi2012cats}, StanfordCars~\cite{krause20133d}, Flowers102~\cite{nilsback2008automated}, Food101~\cite{bossard2014food}, Aircraft~\cite{maji2013fine}, SUN397~\cite{xiao2010sun}, DTD~\cite{cimpoi2014describing}, EuroSAT~\cite{helber2019eurosat}, UCF101~\cite{soomro2012ucf101}). For the $1\times 1$ collage prompt, we referenced the zero-shot experiment results from GPT4Vis~\cite{wu2023gpt4vis}. Using the API service provided by OpenAI, we evaluated the recognition performance for $2\times2$ and $3\times3$ collage prompts. The specific model version used was ``gpt-4-1106-vision-preview''. We used low-resolution to input images and set a random seed to ensure deterministic results.

\subsection{Benchmark Results of Collage Prompting}

Our analysis reveals that utilizing Collage Prompting with GPT-4V for image recognition significantly reduces inference costs without substantially compromising accuracy as shown in Table \ref{tab:cost_acc}. While the ImageNet-1K dataset presents greater challenges due to long text labels, leading to a more drop in accuracy, accuracy on other medium-sized datasets remains substantial. By employing different configurations of collage prompting, including larger grid sizes ($2\times2$ and $3\times3$), we demonstrated a significant decrease in usage costs—approximately to 1/4 and 1/9 of the cost for single images, respectively. Despite a decrease in Top-1 accuracy as grid size increases, our baseline methods for optimizing these collage arrangements significantly reduce accuracy loss, achieving over 5\% higher accuracy than random arrangements. The baseline approach highlights the balance between cost efficiency and performance preservation, making it a practical solution for leveraging large multi-modal models like GPT-4V in resource-constrained scenarios.

Our analysis demonstrates the effectiveness of collage prompting in enhancing cost-efficiency across various datasets, with $n \times n$ grid collages significantly outperforming single $1\times1$ images in terms of the Cost-Effective Ratio (CER) as shown in Table \ref{tab:cost_acc}. The $3\times3$ grids, optimized through our collage graph optimization method, show the most notable improvements in cost efficiency, especially in datasets with simpler labels and less challenging images. Additionally, our Precision-Cost Efficiency (PCE) analysis underscores the trade-off between cost savings and accuracy loss, highlighting that our optimized $2\times2$ and $3\times3$ grid arrangements achieve substantial cost savings while minimizing accuracy loss as demonstrated in Table \ref{tab:cost_acc}, thereby offering a balanced approach to cost-efficient image recognition with GPT-4V. Overall, these results underscore the practicality and efficiency of collage prompting in leveraging large multi-modal models for image recognition tasks under budget constraints.

\subsection{Cost Analysis}
We analyzed the costs associated with using AWS cloud servers for training and inference of traditional CNN and ViT models on the ImageNet-1k dataset, comparing the expenses with those of collage prompting using GPT-4V. Training ResNet-50 and ViT-B/16 from scratch incurred significant costs (\$2,091.63 and \$3,909.46, respectively). In contrast, leveraging GPT-4V for single $1\times1$ image prediction reduced the cost to \$641.25. Additionally, collage prompting with $2\times2$ or $3\times3$ grid configurations further decreased costs to \$170.27 and \$87.75, respectively. The Cost-Effective Ratio (CER) highlights the stark contrast between collage prompting and traditional models. For example, the $3\times3$ grid configuration has a CER 17 times higher than that of ViT-B/16, showcasing the cost efficiency of collage prompting. 

Notably, while models like ViT and CLIP offer impressive capabilities, they entail certain barriers such as data, training, and computational resources. In contrast, GPT-4V enables zero-shot recognition with minimal setup, offering a significant advantage, especially when using collage prompting. This advantage extends to various visual recognition tasks, further emphasizing the practicality and efficiency of GPT-4V in real-world applications.

\section{Limitation and Future work}
\noindent\textbf{Limitations.} 1) \textit{Accuracy Drop}: While collage prompting significantly reduces costs, it does trade off some recognition performance. Despite challenges with datasets like ImageNet-1K, accuracy on medium-sized datasets remains reasonable. Collage prompt is still applicable for tasks with lower accuracy requirements, such as image or video captioning. Our platform will help researchers enhance collage recognition performance, moving closer to the accuracy of standard prompting. 
2) \textit{Collage Prompt in Other LLMs}:
We also tested collage prompting on other open-source and closed-source multimodal vision-language models (\textit{e.g.}, LLAVA-1.5, Gemini 1.5 Pro) and found that these models performed poorly in visual recognition tasks. These models generated non-existent or incorrect labels, produced repetitive outputs, and failed to recognize images within the collage prompt. Appendix~\ref{app:other_models_failures} provides examples of these failures.

\noindent\textbf{Future Work.}
In this paper, we propose a budget-friendly task of collage prompting for GPT-4V's visual recognition and construct a benchmark for learning to optimize the collage prompt. Future work could explore text prompt optimization, visual prompting techniques to learn adversarial noise perturbations, LCP optimization for multiple arrangement candidates, and few-shot/many-shot methods to improve accuracy. Additionally, we will actively maintain and update CollagePrompt, expanding the baseline library, applying it to other multi-modal foundation models, and extending it to broader visual recognition tasks.

\section*{Acknowledgement}
We acknowledge the National Computational Infrastructure of Australia for providing computational resources via the Sydney NCI Scheme. This work was supported in part by the Australian Research Council under Projects DP240101848 and FT230100549.

\bibliography{custom}

\clearpage
\appendix

\appendix
\section{Impact Statement}
While collage prompting offers significant cost savings, especially for large-scale image recognition, it also introduces potential societal implications. The accuracy drop associated with collage prompting could have adverse effects in fields where precise recognition is paramount, such as medical imaging. These implications underscore the necessity for continued exploration into the reliability and safety of leveraging collage prompting for visual recognition tasks, particularly in fields where accuracy is critical.

\subsection{Datasheet for CollagePrompt}
\label{appendix:datasheet}

Here, we provide a datasheet~\cite{gebru2021datasheets} for documenting and ensuring responsible usage of the CollagePrompt Benchmark.

\noindent\textbf{1. Motivation}

\begin{itemize}
\item \emph{For what purpose was the dataset created?} 
This dataset was created as a benchmark for studying the use of collage prompting to reduce the cost of GPT-4V while maintaining reasonable accuracy. It is intended for training and evaluating learning-based collage prompting optimization algorithms.

\item \emph{Who created the dataset (e.g., which team, research group) and on behalf of which entity (e.g., company, institution, organization)?} 
The dataset was created by the authors of this paper.

\item \emph{Who funded the creation of the dataset?} 
The creation of the dataset was funded by the Australian Research Council under Projects DP210101859 and FT230100549.

\end{itemize}

\noindent\textbf{Composition}

\begin{itemize}
\item \emph{What do the instances that comprise the dataset represent (e.g., documents, photos, people, countries)?} The dataset comprises the original information from CollagePrompt's training and validation sets, as well as GPT-4V's prediction results for the collage prompts.

\item \emph{How many instances are there in total (of each type, if appropriate)?} This dataset includes over 110,000 2x2 and 100, 000 3x3 collage prompts with various arrangements of GPT-4V prediction results.

\item \emph{Does the dataset contain all possible instances or is it a sample (not necessarily random) of instances from a larger set?} We randomly sampled 100,000 images from ImageNet to construct the 2x2 and 3x3 collage prompts.

\item \emph{What data does each instance consist of?} 
Each instance consists of a collage and the corresponding GPT-4V prediction results.

\item \emph{Are relationships between individual instances made explicit?} 
Different arrangements of collage prompts and their prediction results from the same set of images are grouped together.

\item \emph{Are there recommended data splits?} 
Yes, the data splits for the training and validation sets are detailed in the JSON files.

\item \emph{Are there any errors, sources of noise, or redundancies in the dataset?} The number of collage images may exceed the number of collage prompts prediction results. We have pre-cleaned the dataset to remove erroneous and unusable predictions of collage prompts.

\item \emph{Is the dataset self-contained, or does it link to or otherwise rely on external resources (e.g., websites, tweets, other datasets)?} The dataset requires downloading the original ImageNet-1K dataset and the downstream evaluation datasets. Using the JSON files that record the original collage information, the code can construct the collage image dataset, which is then used along with our collage prompts prediction results.

\item \emph{Does the dataset contain data that might be considered confidential (e.g., data that is protected by legal privilege or by doctor-patient confidentiality, data that includes the content of individuals’ non-public communications)?} No.

\item \emph{Does the dataset contain data that, if viewed directly, might be offensive, insulting, threatening, or might otherwise cause anxiety?} No.

\end{itemize}

\noindent\textbf{Collection Process}

\begin{itemize}

\item \emph{How was the data associated with each instance acquired?} 
Each collage is constructed from the image datasets according to the order provided in the JSON file. The constructed collage, along with text prompts and dataset category labels, is then input into GPT-4V to obtain the prediction results for each sub-image in the collage.

\item \emph{What mechanisms or procedures were used to collect the data (e.g., hardware apparatuses or sensors, manual human curation, software programs, software APIs)?} Each collage prompt is formatted and input into the GPT-4V API provided by OpenAI to obtain the corresponding prediction results. These results are then post-processed into a standardized, readable format.

\item \emph{Who was involved in the data collection process (e.g., students, crowd workers, contractors), and how were they compensated (e.g., how much were crowd workers paid)?} The data collection process did not involve any manual labor; only API usage fees were incurred.

\item \emph{Over what timeframe was the data collected?} 
The final version of the CollagePrompt dataset was collected in March 2024.

\end{itemize}

\noindent\textbf{Uses}

\begin{itemize}[leftmargin=*]

\item \emph{Has the dataset been used for any tasks already?} Yes, we have used this dataset to train and evaluate our baseline algorithms for optimizing collage prompts.

\item \emph{Is there a repository that links to any or all papers or systems that use the dataset?} Yes, \url{https://collageprompting.github.io/}.

\end{itemize}

\noindent\textbf{Distribution}

\begin{itemize}

\item \emph{Will the dataset be distributed to third parties outside of the entity (e.g., company, institution, organization) on behalf of which the dataset was created?} Yes, the dataset is publicly available online for anyone to access.

\item \emph{How will the dataset be distributed (e.g., tarball on website, API, GitHub)?} The dataset can be downloaded on GitHub.

\item \emph{Will the dataset be distributed under a copyright or other intellectual property (IP) license, and/or under applicable terms of use (ToU)?} Our dataset is distributed under \href{https://creativecommons.org/licenses/by/4.0}{CC BY 4.0}. All codes on the GitHub repository are distributed under the MIT license. 

\item \emph{Have any third parties imposed IP-based or other restrictions on the data associated with the instances?} No.

\item \emph{Do any export controls or other regulatory restrictions apply to the dataset or to individual instances?} No.

\end{itemize}

\noindent\textbf{Maintenance}

\begin{itemize}

\item \emph{Who will be supporting/hosting/maintaining the dataset?} The authors of this paper are supporting/maintaining the dataset.


\item \emph{Is there an erratum?} No.

\item \emph{Will the dataset be updated (e.g., to correct labeling errors, add new instances, delete instances)?} Please check the dataset web page or GitHub repository for any updates.

\item \emph{If others want to extend/augment/build on/contribute to the dataset, is there a mechanism for them to do so?} Yes, they can use the code provided on our GitHub repository to generate data.

\end{itemize}

\subsection{Data Hosting, Licensing, and Maintenance}
The CollagePrompt Benchmark is licensed under the CC BY 4.4, and the data is hosted on Google Drive. All code used for data collection and developing baseline algorithms is distributed under the MIT license. The documentation and model checkpoints are also available on the GitHub repository. The Collage Prompt website (\url{https://collageprompting.github.io/}) is the central hub for all related information, including any future updates and maintenance.

\section{CollagePrompt Benchmark}
\subsection{Prompt Details}
Our text prompts must include the collage filenames and category labels. After extensive experimentation, we have developed a stable version that ensures GPT-4V outputs the correct JSON format without any unrelated content. We also tested various prompting engineering techniques, but they did not significantly affect prediction accuracy. Our prompts can input single or multiple collages for prediction. When using the API, we set the batch size to 4, which does not significantly differ from predicting individual collages.

\noindent \textbf{Text prompts used for $2\times2$ collage}

\begin{lstlisting}[breaklines=true, breakatwhitespace=true, basicstyle=\small\ttfamily]
2x2_prompt = "I want you to act as an Image Classifier. I will provide you with few 2x2 grid collages and a list of optional categories. Your task is to choose the most relevant category for each of the nine images in the grid. Start with the top-left image of each grid and proceed left to right, then down each row. Assign a number index started with 0 for each image in the grid. Provide the prediction in a dict format for each grid collage, key is the number index, and value is the most relevant category for each image in the grid. The final output is also a dictionary. The key is image name of each grid collage, and the value is the prediction for each grid collage in a dict format. Do not provide explanations for your choices or any additional information just the dictionary of predictions in a JSON format. Only output the predictions in one JSON dictionary. Here is the image([]) and its optional categories([]). You have to choose strictly among the given categories and do not give any predictions that are not in the given category."
\end{lstlisting}
    

\noindent \textbf{Text prompts used for $3\times3$ collage}
\begin{lstlisting}[breaklines=true, breakatwhitespace=true, basicstyle=\small\ttfamily]
3x3_prompt = "I want you to act as an Image Classifier. I will provide you with few 3x3 grid collages and a list of optional categories. Your task is to choose the most relevant category for each of the nine images in the grid. Start with the top-left image of each grid and proceed left to right, then down each row. Assign a number index started with 0 for each image in the grid. Provide the prediction in a dict format for each grid collage, key is the number index, and value is the most relevant category for each image in the grid. The final output is also a dictionary. The key is image name of each grid collage, and the value is the prediction for each grid collage in a dict format. Do not provide explanations for your choices or any additional information just the dictionary of predictions in a JSON format. Only output the predictions in one JSON dictionary. Here is the image([]) and its optional categories([]). You have to choose strictly among the given categories and do not give any predictions that are not in the given category."
\end{lstlisting}

\begin{table}[!h]
  \caption{Statistics of datasets used for evaluating collage prompting.}
  \label{tab:dataset_stat}
  \centering
  \resizebox{0.45\textwidth}{!}{%
  \begin{tabular}{@{}l@{\hspace{0.6cm}}r@{\hspace{0.6cm}}r@{\hspace{0.6cm}}r@{}}
    \toprule
    Datasets & Classes & Samples & Label Tokens\\
    \midrule
    ImageNet-1K & 1,000 & 50,000 & 4,834\\
    Caltech101 &100 & 2,465 & 428 \\
    OxfordPets & 37 & 3,669 & 203\\
    StanfordCars & 106 & 8,041 & 1,814\\
    Flowers102 & 102 & 2,463 & 562\\
    Food101 & 101 & 30,300 & 513\\
    FGVCAircraft & 100 & 3,333 & 565\\
    SUN397 & 397 & 19,850 & 1,859\\
    DTD & 47 & 1,692 & 211\\
    EuroSAT & 10 & 8,100 & 49\\
    UCF101 & 101 & 3,783 & 526\\
  \bottomrule
  \end{tabular}
  }
\end{table}
\subsection{Evaluation Datasets}

We evaluate the performance of collage prompt on ImageNet-1K and 10 other common downstream image recognition datasets.
Table \ref{tab:dataset_stat} presents statistics regarding the number of test samples and label tokens for each dataset. 
Label tokens represent the number of tokens encoded by the GPT-4 tokenizer\footnote{https://platform.openai.com/tokenizer}, providing a measure of the textual labels' complexity for each dataset.

\subsection{Dataset Format}
Our dataset is structured as shown below. The training set begins with a random uniform sampling of images from the ImageNet-1k training dataset, followed by the construction of 2x2 and 3x3 collages according to the JSON files containing collage info in the \texttt{metainfo} directory. The evaluation datasets for all downstream datasets are formatted similarly to ImageNet-1k validation, constructing collages based on the standard data segmentation methods. All evaluation datasets consist of complete validation sets derived from ImageNet-1K.

The directory structure of CollagePrompt dataset is as follow:
{\scriptsize\ttfamily
\dirtree{%
.1 /.
.2 train/\DTcomment{}.
.3 collage/\DTcomment{Collage sets of different size}.
.4 2x2.
.4 3x3.
.3 metainfo/.
.4 imagenet\_train\_2x2\_collage\_info.json
    \DTcomment{25,000 collages}.
.4 imagenet\_train\_2x2\_collage\_pred\_info.pkl
    \DTcomment{110,250 collage prompts}.
.4 imagenet\_train\_3x3\_collage\_info.json
    \DTcomment{11,111 collages}.
.4 imagenet\_train\_3x3\_collage\_pred\_info.pkl
    \DTcomment{102,646 collage prompts}.
.3 subset/.
.4 img
    \DTcomment{Sampled 100,000 images from ImageNet-1K}.
.2 val/.
.3 aircraft.
.3 caltech101.
.3 dtd.
.3 eurosat.
.3 food101.
.3 imagenet/.
.4 collage/
    \DTcomment{Collage sets of different size}.
.5 2x2.
.5 3x3.
.4 metainfo/.
.5 imagenet\_val\_2x2\_collage\_info.json
    \DTcomment{12,500 collages}.
.5 imagenet\_val\_3x3\_collage\_info.json
    \DTcomment{5,555 collages}.
.4 set/.
.5 img
    \DTcomment{Validation set (50,000 images) of ImageNet-1K}.
.3 oxflowers.
.3 oxpets.
.3 stcars.
.3 sun397.
.3 ucf101.
.2 README.md.
}
}
\normalsize

\noindent \textbf{Collage Information JSON}  

The format of all `collage\_info.json' files is consistent. Each file contains the original image name, category label, and position index within the collage for each sub-image. Using this JSON file and our provided code on GitHub, users can construct collage images and evaluate the prediction results of collage prompts from GPT-4V. The contents of a collage information JSON file are shown below:

\begin{lstlisting}[numbers=none, basicstyle=\scriptsize\ttfamily]
{
 'a72bca2a3e.jpeg': {
   '0': {'image': 'ILSVRC2012_val_00024101.JPEG',
         'synset_id': 866,
         'label': 'tractor',
         'index': 0},
   '1': {'image': 'ILSVRC2012_val_00011876.JPEG',
         'synset_id': 480,
         'label': 'automated teller machine',
         'index': 1},
   '2': {'image': 'ILSVRC2012_val_00042300.JPEG',
         'synset_id': 842,
         'label': 'swim trunks / shorts',
         'index': 2},
   '3': {'image': 'ILSVRC2012_val_00034749.JPEG',
         'synset_id': 98,
         'label': 'red-breasted merganser',
         'index': 3}},
 'd150b4fd58.jpeg': {
   '0': {'image': 'ILSVRC2012_val_00008159.JPEG',
         'synset_id': 776,
         'label': 'saxophone',
         'index': 0},
   '1': {'image': 'ILSVRC2012_val_00042315.JPEG',
         'synset_id': 123,
         'label': 'spiny lobster',
         'index': 1},
   '2': {'image': 'ILSVRC2012_val_00017726.JPEG',
...
   '3': {'image': 'ILSVRC2012_val_00016843.JPEG',
         'synset_id': 507,
         'label': 'combination lock',
         'index': 3}},
 ...
}
\end{lstlisting}

\noindent\textbf{Collage Prediction JSON.}
The prediction results of GPT-4V for collage prompts are preprocessed and stored in JSON format for ease of use, and then saved as Pickle files to conserve storage space. Files with the suffix `collage\_pred\_info.pkl' contain the prediction results for each collage prompt, including the original image names and their positions within the collage. Below is an example of the contents of such a file:

\begin{lstlisting}[numbers=none, basicstyle=\scriptsize\ttfamily]
{'0b73a3623d.jpeg': {'ord': [1, 2, 0, 3],
  'pred': [0, 0, 1, 1],
  'ori': ['n07697313_12937.JPEG',
   'n03871628_38116.JPEG',
   'n04493381_61391.JPEG',
   'n02086910_6135.JPEG']},
 'eb46c53c56.jpeg': {'ord': [3, 2, 1, 0],
  'pred': [1, 1, 0, 1],
  'ori': ['n07697313_12937.JPEG',
   'n03871628_38116.JPEG',
   'n04493381_61391.JPEG',
   'n02086910_6135.JPEG']},
 '789e579a78.jpeg': {'ord': [3, 0, 2, 1],
  'pred': [1, 1, 0, 0],
  'ori': ['n07697313_12937.JPEG',
   'n03871628_38116.JPEG',
   'n04493381_61391.JPEG',
   'n02086910_6135.JPEG']},
 'f6810cfeb6.jpeg': {'ord': [0, 2, 3, 1],
  'pred': [1, 0, 1, 0],
  'ori': ['n07697313_12937.JPEG',
   'n03871628_38116.JPEG',
   'n04493381_61391.JPEG',
   'n02086910_6135.JPEG']},
 '2ed874ee56.jpeg': {'ord': [3, 0, 1, 2],
...
  'ori': ['n04335435_13775.JPEG',
   'n04023962_6300.JPEG',
   'n04507155_1825.JPEG',
   'n04447861_2306.JPEG']},
 ...
}
\end{lstlisting}

These files can be used to construct the collage image sets and to train and evaluate baseline methods for optimizing collage prompts. We also provide a complete download link for the collage image sets on GitHub, which can be used to reproduce the experimental results.

\begin{figure*}[!h]
    \centering
    \includegraphics[width=5.5in]{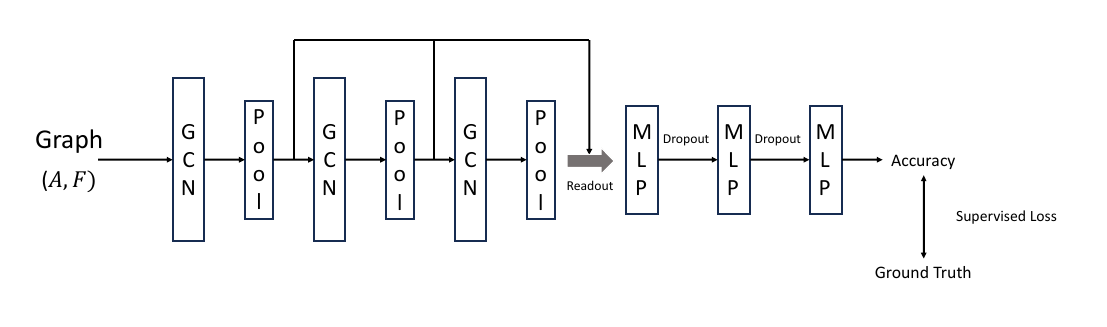}
                \vskip -0.1in
    \caption{Network architecture of collage predictor.}
    \label{fig:gcn_arch}
\end{figure*}
\section{Experimental Details}

\noindent\textbf{Network Details.}
The overall architecture of the collage predictor is depicted in Figure \ref{fig:gcn_arch}. It comprises multiple graph convolutional and pooling layers. The graph convolutional layers aggregate information from neighboring nodes, while the graph pooling layers retain the sub-graph information for each node. This structure preserves the basic graph structural information and facilitates message passing.
By learning graph representations hierarchically and summarizing the node representations in each layer using readout functions, the graph representations are then input into a multi-layer perception (MLP) to perform graph regression prediction tasks, specifically predicting the overall accuracy of the collage graph.

\noindent\textbf{Details of Predictor Training.}
For training the collage predictor, both evaluation datasets for $2\times2$ and $3\times3$ collage prompt are split into training and validation sets at a 9:1 ratio, where 90\% of the data is allocated for training and the remaining 10\% for validation. The collage predictor is trained with a batch size of 512 and a learning rate of 0.001 for 500 epochs. We utilize the Mean Squared Error (MSE) loss function during training.
The node input feature dimension of the collage graph network is set to 512. The network architecture consists of three convolutional layers and pooling layers, with a pooling ratio of 0.5. We employ the Adam optimizer to optimize the model. The dimensions of the three MLP layers are set to [256, 128, 64]. During training, we utilize an early stopping strategy to prevent overfitting. We trained the model for approximately 8 hours using an Nvidia GPU RTX 2080TI.

\noindent\textbf{Details of LCP.}
\label{app:lcp_details}
The pseudocode for our LCP algorithm is provided in Algorithm \ref{app:algorithm_lcp}. When predicting arrangements using LCP, we employ uniform crossover without allowing duplicate genes and random mutation to introduce variation in the predicted arrangements for both $2\times2$ and $3\times3$ collage prompts. For the $3\times3$ collage prompt, we initialize the population with 100 arrangements. In each generation, we select the top 20 arrangements with the highest accuracy to serve as parents for crossover and mutation. The evolution process continues for 10 generations, and we terminate it when the saturation threshold reaches 3. Similarly, for the $2\times2$ collage prompt, we begin with an initial population of 5 arrangements. We retain the top 3 arrangements in each generation based on accuracy for further reproduction. The evolution process runs for 5 generations, and we stop it when the saturation threshold also reaches 3. Finally, we evaluate the predicted best arrangements by feeding them in batches of 4 to the GPT-4V API to obtain the actual prediction accuracy. 
\begin{figure*}[!t]
    \centering
    \includegraphics[width=5.5in]{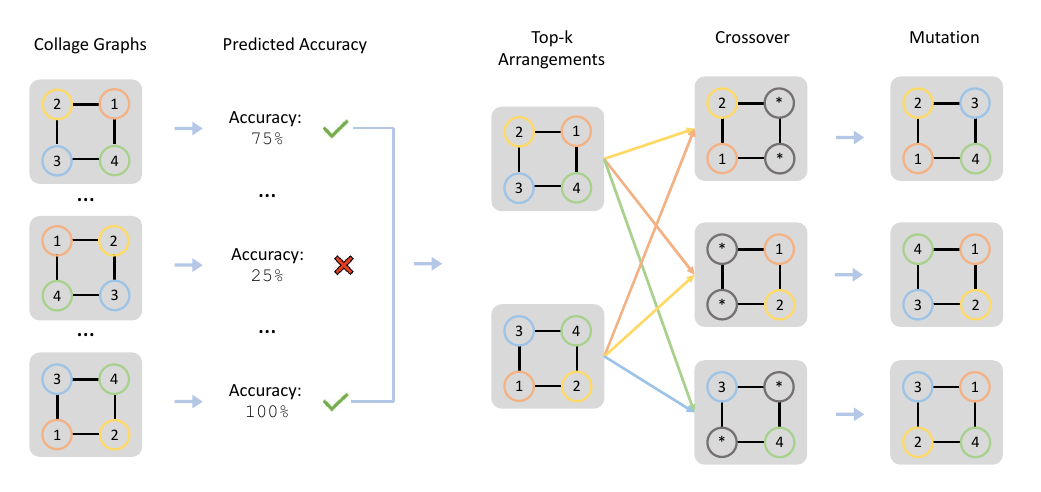}
    \caption{The process of Crossover and Mutation in the proposed LCP.}
    \label{fig:cross_mute_vis}
\end{figure*}

\begin{table}
  \caption{Top-1 accuracy, inference time and cost of collage prompts with different number of images $K$ in GPT-4V's image recognition.
  }
  \label{tab:acc_diff_collage_sizes}
  \centering
  \begin{tabular}{l@{\hspace{0.35cm}}r@{\hspace{0.35cm}}r@{\hspace{0.35cm}}r}
    \toprule
    $K$  & Top-1 Acc & Time & Cost \\
    \midrule
    1$\times$1 & 62.0\% & 8.15s & \$51.30 \\
    2$\times$2 & 39.4\% & 2.75s &  \$12.83  \\
    3$\times$3 & 28.1\% & 1.34s & \$5.70  \\
    4$\times$4 & 21.5\%  & 1.05s & \$3.21  \\
    5$\times$5 & 11.9\% & 0.95s & \$2.05  \\
  \bottomrule
  \end{tabular}
\vskip -0.2in
\end{table}

\begin{algorithm}[t]
  \SetAlgoNoLine
  \textbf{Parameters:} $n$: size of arrangement candidates, $m$: size of selected arrangement, $\chi$: crossover rate, $\mu$: mutation rate\;

  \textbf{Initialise generation 0:}\;
  
  $k := 0$\;
  
  $P_k$ := a set of $n$ randomly-generated arrangements\;
  
   \textbf{Evaluate} $P_k$:\;
  
  Compute $fitness(i)$ for each $i \in P_k$\;
  
  \While{\textbf{not} stop-criterion}{

    // \textbf{Create generation $k + 1$:}\;

    // \textbf{1. Copy:}\; 
    
    Select Top-$m$ arrangements from $P_k$; insert into $P_{k+1}$\;
    
    // \textbf{2. Crossover:}\;
    
    Randomly pop out two arrangements from Top-$m$; pair them up; produce $\chi \times n$ new arrangements; insert the arrangements into $P_{k+1}$\;
    
    // \textbf{3. Mutate:}\;
    
    Select $\mu \times n$ arrangements of $P_{k+1}$; invert a randomly-selected bit in each\;
    
    // \textbf{Evaluate $P_{k+1}$:}\;
    
    Compute fitness$(i)$ for each $i \in P_{k+1}$\;
    
    // \textbf{Increment:}\;
    $k := k + 1$\;
    
  }
  \Return the fittest arrangement from $P_k$\;
  \caption{LCP Algorithm for Collage Arrangement Optimization}
  \label{app:algorithm_lcp}
\end{algorithm}

\begin{figure*}[!tbp]
    \centering
    \begin{minipage}[b]{1\linewidth}
        \begin{subfigure}[b]{0.22\textwidth}
            \centering
            \includegraphics[width=0.8\textwidth]{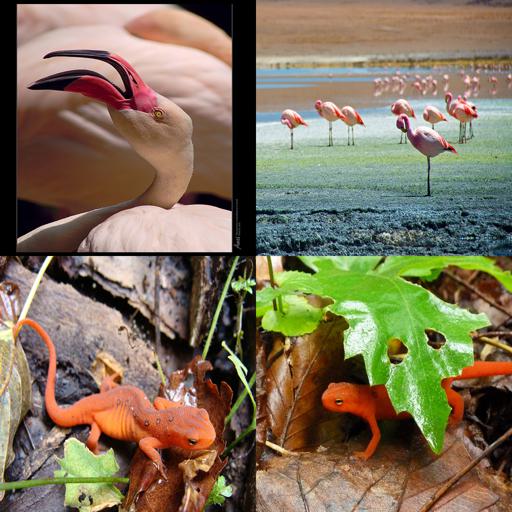}
            \caption{"0: ‘flamingo’, 1: ‘flamingo’, 2: ‘eft’, 3: ‘eft’"}
            \label{fig:ob2:c1a}
        \end{subfigure}
        \hfill
        \begin{subfigure}[b]{0.22\textwidth}
            \centering
            \includegraphics[width=0.8\textwidth]{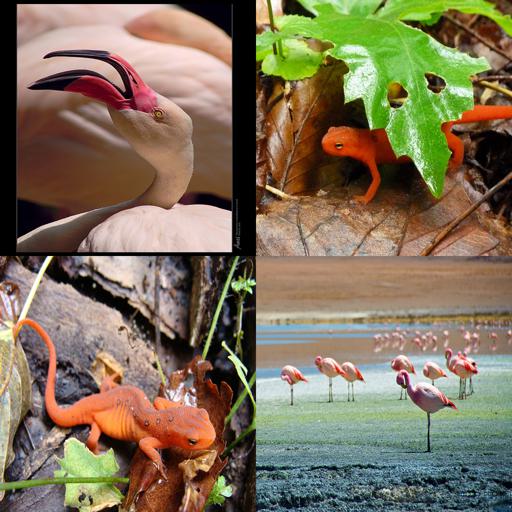}
            \caption{"0: ‘flamingo’, 1: ‘eft’, 2: ‘{\color{red}flamingo}’, 3: ‘flamingo’"}
            \label{fig:ob2:c1b}
        \end{subfigure}
        \hfill
        \begin{subfigure}[b]{0.22\textwidth}
            \centering
            \includegraphics[width=0.8\textwidth]{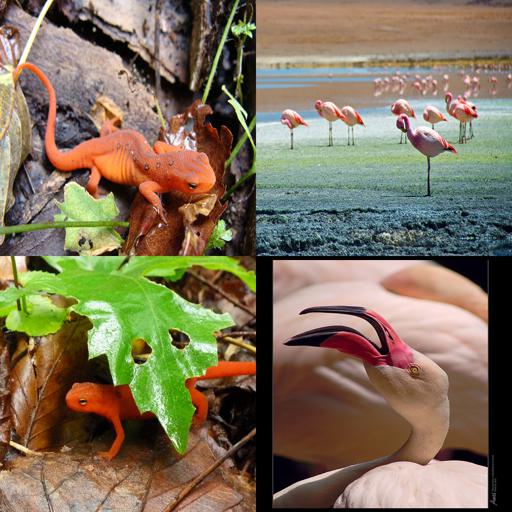}
            \caption{"0: ‘eft’, 1: ‘flamingo’, 2: ‘eft’, 3: ‘flamingo’"}
            \label{fig:ob2:c1c}
        \end{subfigure}
        \hfill
        \begin{subfigure}[b]{0.22\textwidth}
            \centering
            \includegraphics[width=0.8\textwidth]{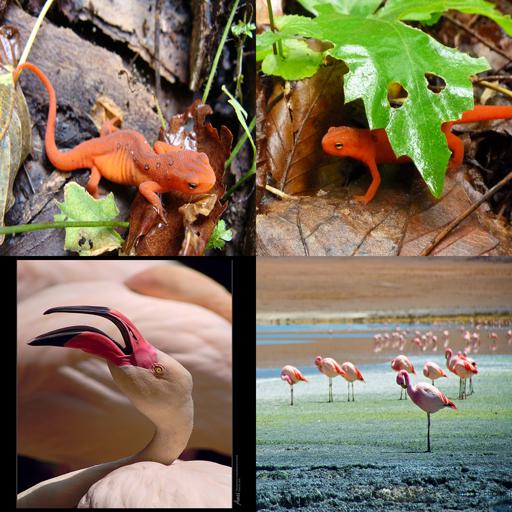}
            \caption{"0: ‘eft’, 1: ‘eft’, 2: ‘flamingo’, 3: ‘flamingo’"}
            \label{fig:ob2:c1d}
        \end{subfigure}
        \begin{subfigure}[b]{0.22\textwidth}
            \centering
            \includegraphics[width=0.8\textwidth]{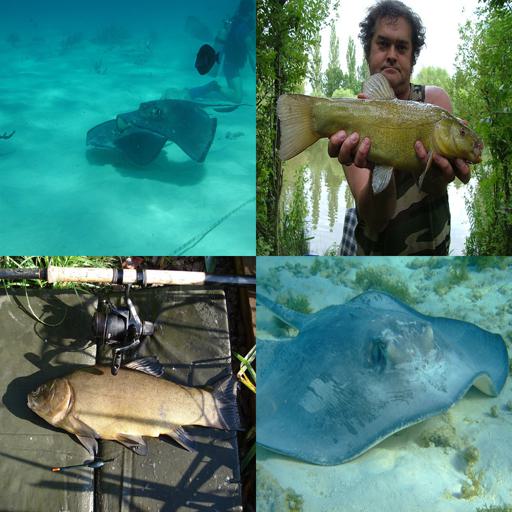}
            \caption{"0: ‘stingray’, 1: ‘tench’, 2: ‘tench’, 3: ‘{\color{red}electric ray}’"}
            \label{fig:ob2:c1a}
        \end{subfigure}
        \hfill
        \begin{subfigure}[b]{0.22\textwidth}
            \centering
            \includegraphics[width=0.8\textwidth]{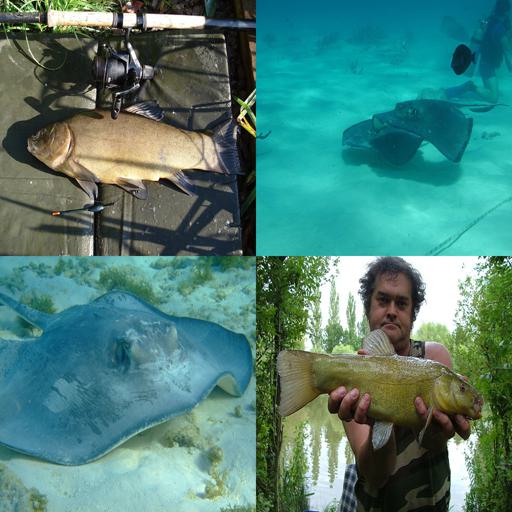}
            \caption{"0: ‘tench’, 1: ‘stingray’, 2: ‘{\color{red}tench}’, 3: ‘tench’"}
            \label{fig:ob2:c1b}
        \end{subfigure}
        \hfill
        \begin{subfigure}[b]{0.22\textwidth}
            \centering
            \includegraphics[width=0.8\textwidth]{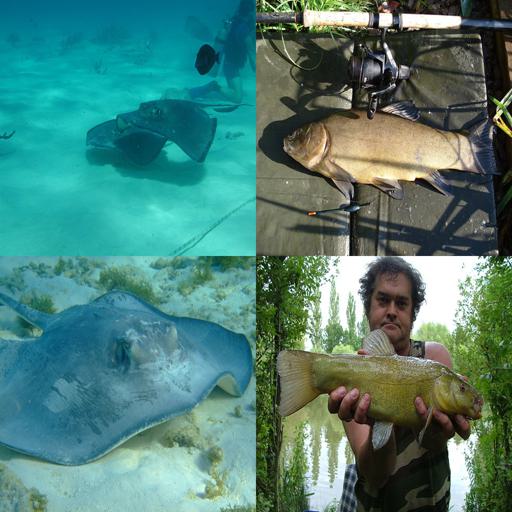}
            \caption{"0: ‘stingray’, 1: ‘tench’, 2: ‘stingray’, 3: ‘tench’"}
            \label{fig:ob2:c1c}
        \end{subfigure}
        \hfill
        \begin{subfigure}[b]{0.22\textwidth}
            \centering
            \includegraphics[width=0.8\textwidth]{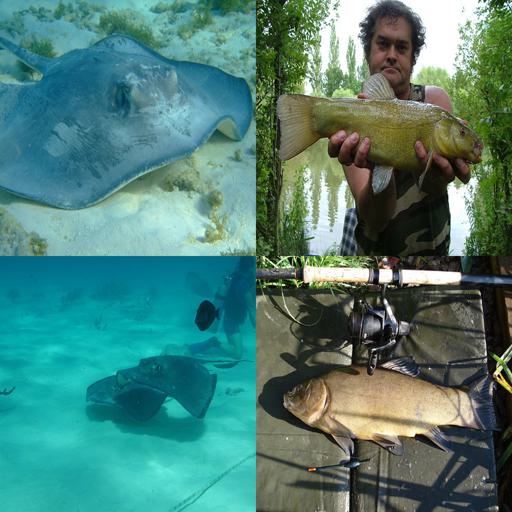}
            \caption{"0: ‘stingray’, 1: ‘tench’, 2: ‘stingray’, 3: ‘tench’"}
            \label{fig:ob2:c1d}
        \end{subfigure}
        
        \begin{subfigure}[b]{0.22\textwidth}
            \centering
            \includegraphics[width=0.8\textwidth]{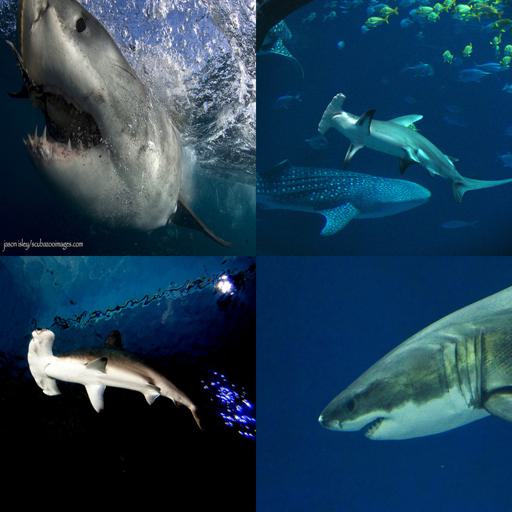}
            \caption{"0: ‘great white shark’, 1: ‘{\color{red}great white shark}’, 2: ‘hammerhead shark’, 3: ‘great white shark’"}
            \label{fig:ob2:c1a}
        \end{subfigure}
        \hfill
        \begin{subfigure}[b]{0.22\textwidth}
            \centering
            \includegraphics[width=0.8\textwidth]{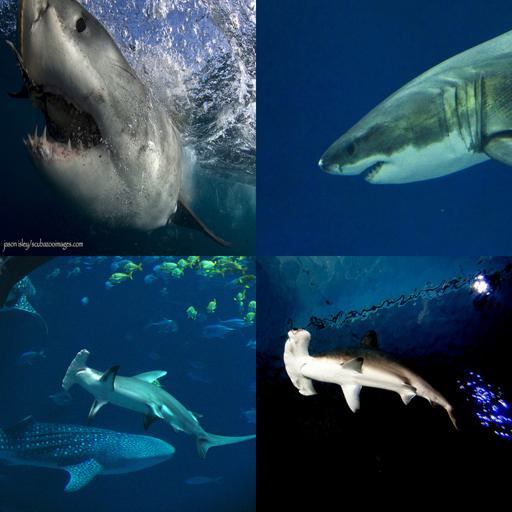}
            \caption{"0: ‘great white shark’, 1: ‘great white shark’, 2: ‘hammerhead shark’, 3: ‘hammerhead shark’"}
            \label{fig:ob2:c1b}
        \end{subfigure}
        \hfill
        \begin{subfigure}[b]{0.22\textwidth}
            \centering
            \includegraphics[width=0.8\textwidth]{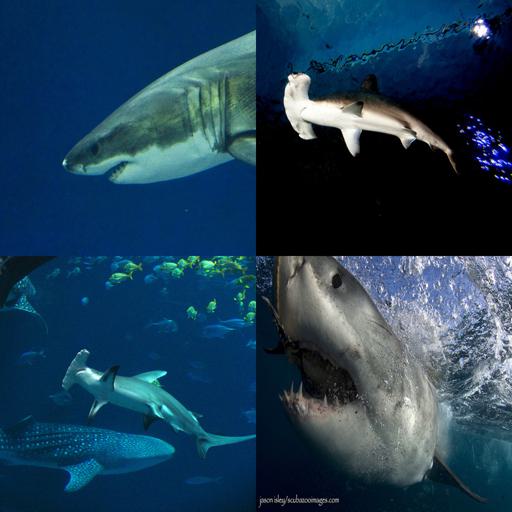}
            \caption{"0: ‘great white shark’, 1: ‘{\color{red}great white shark}’, 2: ‘{\color{red}great white shark}’, 3: ‘great white shark’"}
            \label{fig:ob2:c1c}
        \end{subfigure}
        \hfill
        \begin{subfigure}[b]{0.22\textwidth}
            \centering
            \includegraphics[width=0.8\textwidth]{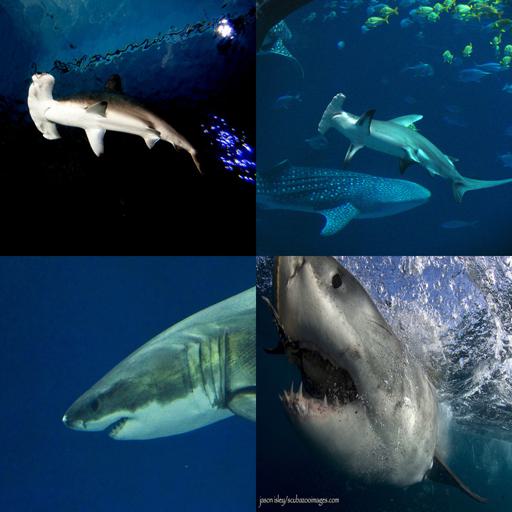}
            \caption{"0: ‘hammerhead shark’, 1: ‘hammerhead shark’, 2: ‘great white shark’, 3: ‘great white shark’"}
            \label{fig:ob2:c1d}
        \end{subfigure}
        \caption{Examples of \textbf{Category Clustering}, showing GPT-4V's predictions for images of the same category placed adjacently or non-adjacently.
        }
        \label{fig:obs2_cases}
    \end{minipage}
\end{figure*}
\noindent\textbf{Details of Crossover and Mutation.}
During the iterative process of optimizing arrangements using LCP, crossover and mutation of arrangements are involved. The specific processes of crossover and mutation are illustrated in Figure \ref{fig:cross_mute_vis}.
At the initial stage of each iteration, our LCP algorithm predicts the accuracy of each initial arrangement using the collage predictor and retains the top-k collage arrangements. Then, any two arrangements from the top-k are randomly selected for node crossover to obtain n partial initial node arrangements for the collages. Finally, the remaining nodes are randomly allocated (mutated) to the blank positions in the collages.

\noindent\textbf{Alternative Optimization Methods.} 
Collage optimization is a discrete problem, and our initial exploration of various methods revealed that gradient-based approaches required costly gradient estimation, making training difficult. We opted for the genetic algorithm due to its simplicity and efficiency in searching for optimal collage arrangements. To support further research, we provide a publicly available benchmark platform to encourage the development of advanced algorithms that enhance cost-efficiency for GPT-4V and similar models, fostering improvements in both performance and cost reduction for large-scale multimodal AI systems.

\section{Visualized Results and Analysis} 
\label{app:visual_analysis}

\begin{figure*}[!t]
    \centering
    \includegraphics[width=5.5in]{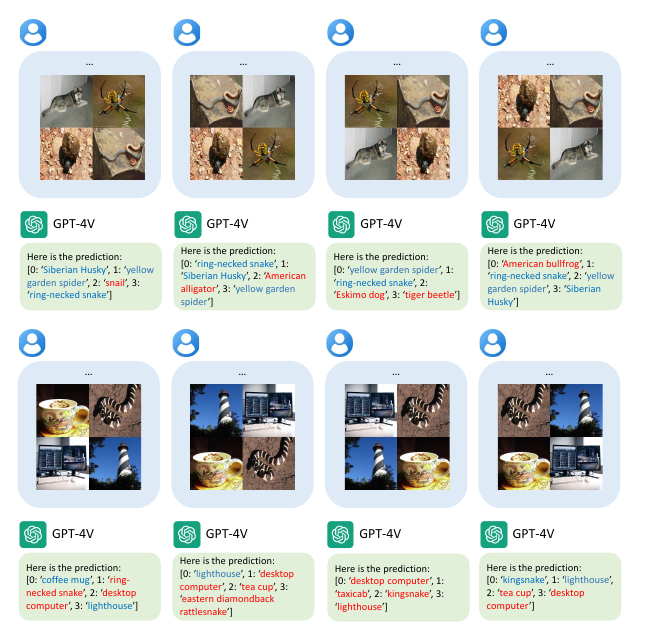}
    \caption{
    Examples of \textbf{Localization Errors}: Two cases that demonstrate different arrangements within the collage prompt lead to different accuracy of classification. \textcolor{blue}{Blue} indicates an accurate prediction while \textcolor{red}{red} indicates a wrong prediction.
}
\vskip -0.17in
    \label{fig:app_case_1}
\end{figure*}

\begin{figure*}[!t]
    \centering
    \includegraphics[width=5.5in]{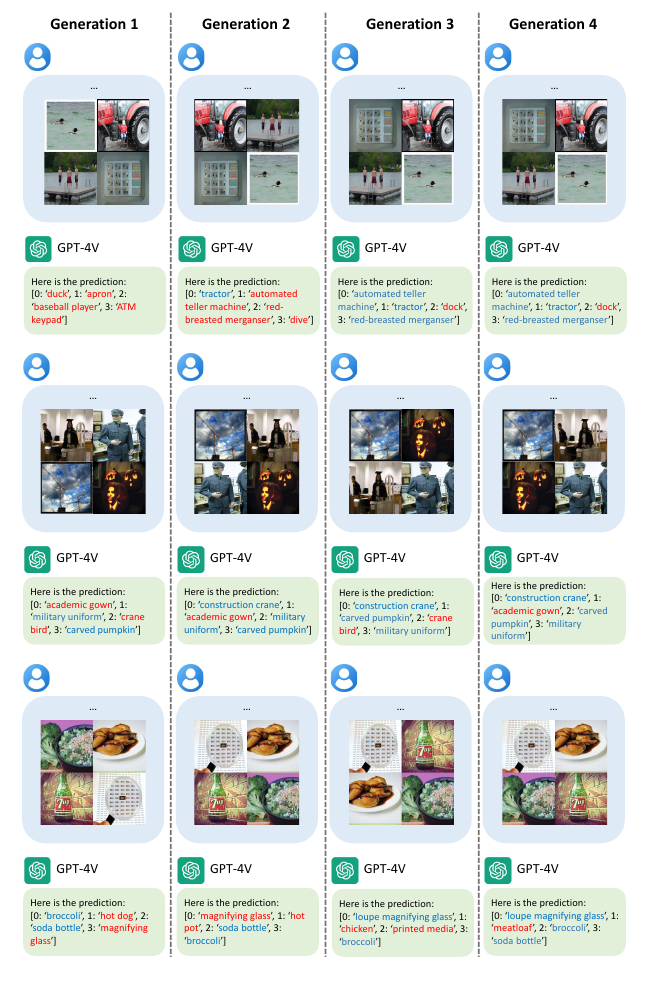}
    \caption{Illustration of optimized collage arrangements and corresponding GPT-4V predictions across different generations, generated using the LCP algorithm.}
    \label{fig:alm_case}
\end{figure*}
\subsection{More cases about Category Clustering}
As shown in Figure \ref{fig:obs2_cases}, we provide three examples of category clustering, illustrating how GPT-4V's predictions are influenced by the adjacency of images within the same category.

In the first row, we observe the behavior of GPT-4V when identifying flamingo and eft. In subfigure (a), where flamingos are grouped together and efts are grouped together, the predictions are accurate with both images correctly identified as flamingo and eft. However, in subfigure (b), when an eft is placed diagonally and not grouped with other efts, GPT-4V incorrectly predicts eft as flamingo in one instance. When flamingos and efts are grouped together again in subfigures (c) and (d), the predictions return to being correct.

The second row demonstrates the prediction tendencies for tench and stingray. In subfigure (a), when tenches and stingrays are grouped together, GPT-4V accurately predicts their respective categories. However, in subfigure (b), with tenches and stingrays positioned diagonally and not grouped together, GPT-4V incorrectly predicts ‘stingray’ as ‘tench’. This misclassification persists in subfigures (c) and (d) when the tenches and stingrays are diagonally positioned, highlighting the impact of image arrangement on GPT-4V's predictions.

In the third row, we observe the interactions between great white shark and hammerhead shark. In subfigures (a), (b), and (c), where the sharks are positioned diagonally and not grouped together, GPT-4V consistently misclassifies hammerhead shark as ‘great white shark’. However, in subfigure (d), when the sharks are grouped together, GPT-4V accurately distinguishes between the two shark species. These examples underscore the importance of grouping images of the same category together in collage prompting to improve GPT-4V's accuracy.

These observations highlight the significance of image arrangement in collage prompting. When images of similar categories are positioned adjacently, GPT-4V's accuracy improves. Conversely, non-adjacent placement, especially diagonal positioning, increases the likelihood of misclassification. This underscores the need for careful consideration of image layout in tasks requiring high recognition accuracy, as the arrangement can substantially impact the performance of visual recognition models.

\begin{table*}[t]
  \caption{Baseline method v.s. Brute Force Solution. $3\times3$ Grid Collage.}
  \vskip 0.05in
  \label{tab:ga_vs_bf}
  \centering
    \setlength{\tabcolsep}{5mm}{
  \begin{tabular}{l|c|cccc@{}}
    \toprule
    K & Method & Steps & Time & Fitness & Accuracy \\
    \midrule
     \multirow{8}{*}{\begin{turn}{90}\textbf{3}$\times$\textbf{3}\end{turn}}
     & \multirow{4}{*}{LCP (Baseline)} & 100 & 1.631 & 0.114 & 0.330 \\
     & & 500 & 4.479 & 0.138 & 0.334 \\
     &  & 1000 & 7.727 & 0.141 & 0.330 \\
     &  & 1500 & 10.952 & 0.141 & 0.334 \\
    \cmidrule{2-6}
    &\multirow{4}{*}{Brute Force} & 100 & 0.670 & 0.102 & 0.330 \\
    & & 500 & 3.301 & 0.124 & 0.329 \\
    & & 1000 & 6.812 & 0.131 & 0.326 \\
    & & 1500 & 9.856 & 0.136 & 0.328 \\
    \midrule
     \multirow{8}{*}{\begin{turn}{90}\textbf{2}$\times$\textbf{2}\end{turn}}    
    &\multirow{4}{*}{LCP (Baseline)} & 5 & 0.078 & 0.0419 & 0.445 \\
    & & 10 & 0.107 & 0.0439 & 0.446 \\
    & & 15 & 0.131 & 0.0433 & 0.448 \\
    & & 24 & 0.187 & 0.0439 & 0.445 \\
    \cmidrule{2-6}    
    &\multirow{4}{*}{Brute Force} & 5 & 0.033 & 0.040 & 0.449 \\
    & & 10 & 0.066 & 0.0428 & 0.448 \\
    &  & 15 & 0.099 & 0.0433 & 0.448 \\
    & & 24 & 0.164 & 0.0436 & 0.451 \\
  \bottomrule
  \end{tabular}
  }
\end{table*}

\begin{figure*}[!t]
    \centering
    \includegraphics[width=5.5in]{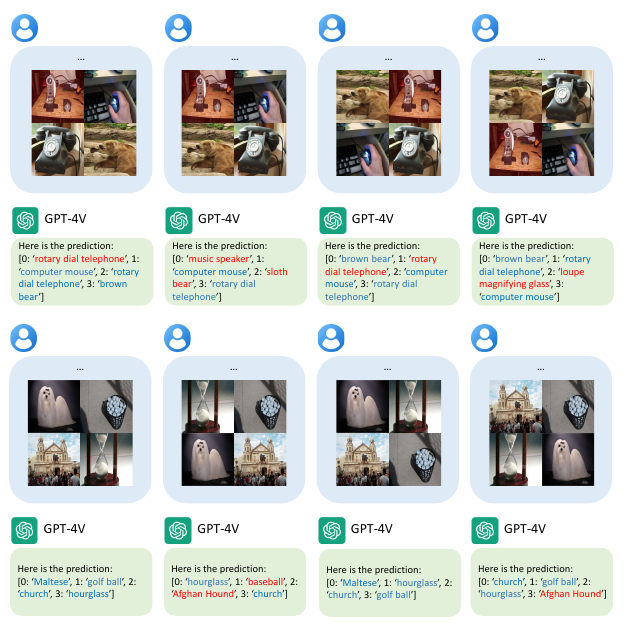}
    \caption{
    Examples of \textbf{Localization Errors}: Two cases that demonstrate different grids within the collage prompt lead to different accuracy of classification.
    \textcolor{blue}{Blue} indicates an accurate prediction while \textcolor{red}{red} indicates a wrong prediction.}
    \label{fig:app_case_2}
\end{figure*}
\subsection{More cases about Localization Errors}
The visualization of collage prompt reveals distinct variations in the recognition accuracy of collage images by GPT-4V across different positions within the collage. Specifically, images positioned in the top-left corner exhibit the highest recognition accuracy, while those in the bottom-left corner demonstrate the lowest accuracy. For instance, in the first row of Figure \ref{fig:app_case_1}, the ``Siberian Husky'' is misclassified when positioned in the bottom-left corner but correctly identified in other positions. Moreover, relocating challenging samples to the top-left corner notably enhances GPT-4V's identification accuracy. For instance, in the second row of Figure \ref{fig:app_case_1}, the ``coffee mug'', identified as a challenging sample, is correctly recognized only when placed in the top-left corner, whereas it is misclassified in other positions. Similarly, such phenomena are observed in the second row of Figure \ref{fig:app_case_2}.

Additionally, we observed instances of mislocalization during collage image recognition by GPT-4V. This phenomenon entails the correct label of an image within the collage being predicted for the adjacent image's position. For example, in the second row of Figure \ref{fig:app_case_1}, the ``lighthouse'' positioned in the bottom-left corner of the third collage is misclassified as the last image in the bottom-right corner. This mislocalization is more pronounced in the first row of Figure \ref{fig:app_case_2}, where the ``projector'' is consistently misclassified as a ``rotary dial telephone'' when adjacent, but correctly classified as other categories when positioned diagonally. This observation offers insight into why the recognition accuracy of images in collages, particularly in datasets like EuroSAT, surpasses that of single images. When images of the same category are juxtaposed in a collage, they provide mutual cues for GPT-4V to predict the correct labels. This phenomenon was further validated through experimentation.
These findings underscore the importance of considering the spatial arrangement of images within a collage when interpreting recognition accuracy and offer insights into the mechanisms underlying GPT-4V's recognition performance in such contexts.

\subsection{Arrangement Optimization in LCP}
Figure \ref{fig:alm_case} displays various optimal collage arrangements and their corresponding predictions by GPT-4V across different generations, as generated by the LCP algorithm. In the first row examples, the ``automated teller machine'' was initially mispredicted in the first two generations but was correctly placed in the top-left corner in the third generation, resulting in a correct prediction by GPT-4V. The optimal arrangement remained consistent in the fourth generation, suggesting that the LCP algorithm stabilized after achieving the best arrangement.

In the second row examples, the ``construction crane'' was mispredicted when placed in the bottom-left corner in the first generation. However, it was correctly positioned in the top-left corner in the second generation and remained there in subsequent iterations. This indicates that the LCP algorithm learned to place challenging samples in the top-left corner for improved prediction accuracy, while simpler samples were positioned in the bottom-left corner to enhance overall collage recognition accuracy.

In the third row examples, the ``loupe magnifying glass'' was initially placed in the bottom-right corner in the first generation, resulting in a misprediction by GPT-4V. Subsequently, in the second generation, the LCP algorithm positioned it in the top-left corner, still leading to a misprediction. However, in the following iterations, ``loupe magnifying glass'' persisted in the top-left corner, indicating the LCP predictor's confidence in this arrangement despite the initial misprediction. Eventually, in the later generations, the correct prediction was made when the ``loupe magnifying glass'' was placed in the top-left corner again. This example highlights the robustness of our trained LCP predictor and suggests some stochasticity in the prediction outcomes of GPT-4V.

These cases in Figure \ref{fig:alm_case} further demonstrate that GPT-4V's accuracy in recognizing images within a collage varies across different positions. The LCP algorithm successfully learns the positions that yield the highest and lowest accuracy and optimally arranges the images to enhance the overall collage recognition accuracy by GPT-4V.

\section{Ablation Study}

\subsection{Cost-Efficiency Analysis of Collage Sizes}

Table \ref{tab:acc_diff_collage_sizes} illustrates the Top-1 accuracy of collage sizes ranging from $1\times1$ to $5\times5$ random grid arrangements. It also presents the inference time per image and the associated cost of using the GPT-4V API for inference per 1000 images. Notably, transitioning from single $1\times1$ images to $2\times2$ grid collages results in a reduction in accuracy of approximately 22.6\%. However, the inference time and API usage cost decrease by nearly fourfold. Subsequently, each increment in grid size, from $2\times2$ to $5\times5$, leads to a decrement in accuracy by nearly 10\%. Given the impracticality of using $4\times4$ and $5\times5$ grid sizes due to their significantly lower accuracy and the extensive search space for grid arrangements, focusing on optimizing the arrangement learning solely for $2\times2$ and $3\times3$ grids holds practical value. This is because $2\times2$ and $3\times3$ grids maintain acceptable accuracy levels while ensuring sufficiently low costs.

\subsection{Comparison of Optimization Methods}

Table \ref{tab:ga_vs_bf} compares the efficacy of random initialization, brute force search using a trained model predictor, and optimization using our LCP algorithm for obtaining optimal grid arrangements for both $2\times2$ and $3\times3$ collage sizes. It is evident from the table that grid arrangements obtained through the model predictor-based search and optimization generally outperform those obtained through random initialization.

Moreover, the grid arrangements optimized using the LCP algorithm demonstrate higher accuracy levels while requiring fewer steps and less time compared to brute force search. For instance, the LCP algorithm for $3\times3$ grids achieves a similar accuracy as brute force search but with only 500 steps, three times fewer than brute force, and in half the time. This emphasizes the efficiency and effectiveness of using LCP algorithms for optimizing grid arrangements.

\section{Failure Cases from Other LVLMs}
\label{app:other_models_failures}

To illustrate the challenges that other large vision-language models (LVLMs) face in handling collage prompts, we present several failed response examples from different models. These outputs highlight the significant limitations of existing non-GPT-4V models in visual recognition tasks involving collage prompts.

\noindent \textbf{Qwen2-VL-7B-Instruct}

\begin{lstlisting}[breaklines=true, breakatwhitespace=true, basicstyle=\small\ttfamily]
https://red-brew
concertedietet stimme in to the one of the the the the the the the the the the the the the the the the the the the the the the the the the the the the the the the the the the the the the the the the the the the the the the the the the the the the the the the the the the the the the the the the the the the the the the the the the
\end{lstlisting}

\noindent \textbf{LLAVA-v1.6-vicuna-7b}

\begin{lstlisting}[breaklines=true, breakatwhitespace=true, basicstyle=\small\ttfamily]
{
  "79f1ff56be.jpeg": {
    "airplane": 0,
    "bird": 0,
    "car": 0,
    "cat": 0,
    "deer": 0,
    "dog": 0,
    "horse": 0,
    "monkey": 0,
    "ship": 0,
    "truck": 0
  }
}
\end{lstlisting}

\noindent \textbf{InternVL2-8B}

\begin{lstlisting}[breaklines=true, breakatwhitespace=true, basicstyle=\small\ttfamily]
{
  "79f1ff56be_topleft.jpg": {
    __caption_uuidf5Gddf4 vardr steht in the top left of the assigned class for this picture is the wrong one on the integration of the - at the in the of the of the of the of the of the of the of the of the of the of the of the of the of the of the of the of the of the of the of the of the of the
  }
}
\end{lstlisting}

\noindent \textbf{Gemini Pro 1.5}

\begin{lstlisting}[breaklines=true, breakatwhitespace=true, basicstyle=\small\ttfamily]
Sorry I can't help with that image. Try uploading another image or describing the image you tried to upload, and I can help you that way.
\end{lstlisting}

These examples demonstrate that current non-GPT-4V models struggle significantly with collage prompts, producing outputs that are either nonsensical, repetitive, or completely uninformative. These results underscore the infeasibility of benchmarking these models against GPT-4V for collage-based visual recognition tasks. Future iterations of this work will continue evaluating LVLMs as they evolve to determine if improvements in handling collage prompts emerge.

\end{document}